\PassOptionsToPackage{numbers,sort&compress}{natbib}
\documentclass[acmlarge,screen,nonacm]{acmart}

\usepackage{amssymb}
\usepackage{graphicx}%
\usepackage{bbding}%
\usepackage{pifont}%
\usepackage{multicol}
\usepackage{nicematrix} %
\usepackage{tikz}

\usepackage{fontawesome}

\AtBeginDocument{%
  }

\acmJournal{IMWUT}
\acmVolume{0}
\acmNumber{0}
\acmArticle{0}
\acmMonth{0}

\setcopyright{acmlicensed}
\copyrightyear{2026}
\acmYear{2026}
\acmDOI{XXXXXXX.XXXXXXX}

\acmISBN{978-1-4503-XXXX-X/18/06}

\acmSubmissionID{8717}

\usepackage{booktabs} %
\usepackage{graphicx} %
\usepackage{lipsum}   %
\usepackage{amsmath,amssymb,amsfonts}
\usepackage[noend]{algorithmic}
\usepackage{algorithm}
\renewcommand{\algorithmicrequire}{}  
\usepackage{float}
\usepackage{subfig}
\usepackage{multirow} 
\usepackage{textcomp} 
\usepackage{setspace}
\usepackage{bm}

\usepackage{makecell}

\usepackage{booktabs}
\usepackage{tabularx}
\usepackage{array}
\usepackage{makecell}
\usepackage{pifont}
\usepackage{xcolor}
\usepackage{colortbl}

\definecolor{ReCALMShade}{HTML}{EBF3FA}
\definecolor{CheckColor}{HTML}{21785C}
\definecolor{CrossColor}{HTML}{8B929B}

\usepackage{enumitem}
\usepackage{mathtools,bm}
\usepackage{nicefrac,microtype}
\usepackage{colortbl,multirow,nicematrix}
\usepackage{float,algorithm,algorithmic,setspace}
\usepackage{subfig}
\usepackage{tabularx}
\usepackage{wrapfig}
\renewcommand{\algorithmicrequire}{}
\definecolor{mygreen}{RGB}{37,116,72}
\definecolor{myblue}{RGB}{17,74,150}
\definecolor{my1}{RGB}{0,112,204}
\definecolor{my2}{RGB}{0,128,121}
\definecolor{myred}{RGB}{192,14,14}
\definecolor{myred2}{RGB}{5,5,5}
\definecolor{myred3}{RGB}{0,0,0}
\definecolor{mygray}{RGB}{141,141,141}
\newcommand{\myblueA}[1]{\textcolor{myblue}{#1}}

\newcommand{\myredA}[1]{\textcolor{myred}{#1}}

\newcommand{\answerTODO}[1][]{\textcolor{red}{\bf [TODO]}}
\newcommand{\justificationTODO}[1][]{\textcolor{red}{\bf [TODO]}}

\usepackage{xurl}
\hypersetup{pdfauthor={},pdfsubject={}}

\begin{document}

\title[ReCALM: A Residual-enhanced Crowdsourced Analytic Learning Mechanism for Distributed Continual Service Construction]{ReCALM: A Residual-enhanced Crowdsourced Analytic Learning Mechanism for Distributed Continual Service Construction}

\author{Kejia Fan}
\authornote{Work done as an intern at Peking University. Kejia Fan is also affiliated with Central South University. }
\affiliation{
  \institution{Peking University}
  \country{China}
}

\author{Yajiang Huang}
\affiliation{
  \institution{Central South University}
  \country{China}
}

\author{Jianheng Tang}
\affiliation{
  \institution{Peking University}
  \country{China}
}

\author{Jinfeng Xu}
\affiliation{%
  \institution{University of British Columbia}
  \country{Canada}
}

\author{Huiping Zhuang}
\affiliation{%
  \institution{South China University of Technology}
  \country{China}
}

\author{Anfeng Liu}
\affiliation{
  \institution{Central South University}
  \country{China}
}

\author{Houbing Herbert Song}
\affiliation{
  \institution{University of Maryland, Baltimore County}
  \country{USA}
}

\author{Yunhuai Liu}
\affiliation{
  \institution{Peking University}
  \city{Beijing}
  \country{China}
}

\renewcommand{\shortauthors}{ }

\begin{abstract}
Crowdsourcing has emerged as a promising paradigm for constructing AI services by harnessing the collective capabilities of distributed workers.
As consumers' demands continually evolve, service models must acquire new knowledge while retaining previously learned information.
Centralized approaches require workers to upload private raw data, whereas distributed approaches can improve privacy by avoiding direct raw data transmission.
However, due to the inherited reliance on gradients for backpropagation, existing approaches typically suffer from catastrophic forgetting during continual model updates on heterogeneous data.
Recent studies have introduced analytic learning, a gradient-free paradigm that achieves invariance through recursive closed-form solutions.
Nevertheless, existing analytic methods are typically constrained by single-layer linear networks, which tend to suffer from underfitting due to their inability to perform deep representation learning.
To address this issue, we propose \textbf{ReCALM}, a \textbf{Re}sidual-enhanced \textbf{C}rowdsourced \textbf{A}nalytic \textbf{L}earning \textbf{M}echanism that achieves deep nonlinear invariance for distributed continual service construction.
Inspired by the success of ResNet in gradient-based networks, we develop gradient-free residual blocks for ReCALM with closed-form solutions.
To implement distributed continual training for ReCALM, we further devise a cross-recursive strategy that incorporates both layer-wise and phase-wise recursion via least squares.
Both theoretical and experimental analyses verify ReCALM’s dual superiority in terms of its heterogeneity invariance and deep representation.
With 20 residual layers, ReCALM improves accuracy by 8.85\%–9.83\% over the single-layer analytic baseline, while reducing computational costs by over 94\% compared with gradient-based baselines.
\end{abstract}


\maketitle

\section{Introduction}

Mobile crowdsensing is a promising paradigm for ubiquitous AI service construction by leveraging the collective sensing capabilities of distributed workers~\cite{guo2015mobile,wang2021csmc,zhao2024crowdgeo}.
In a crowdsourcing system, consumers express service demands, the platform (also referred to as the server) constructs AI models, and workers (also referred to as clients) contribute either locally collected data or derived statistics with embedded knowledge~\cite{CALM}.
As consumer demands evolve, the platform (in collaboration with workers) must continually update its service models to incorporate new knowledge while preserving previously acquired knowledge~\cite{FCL-0, FCL-8-FedWeIT}.
For instance, an image-recognition service may need to progressively expand from recognizing animals to plants and vehicles without compromising its existing capabilities.
This process is referred to as crowdsourced continual service construction.

\begin{figure}[t]
    \centering
    \includegraphics[width=0.80\linewidth]{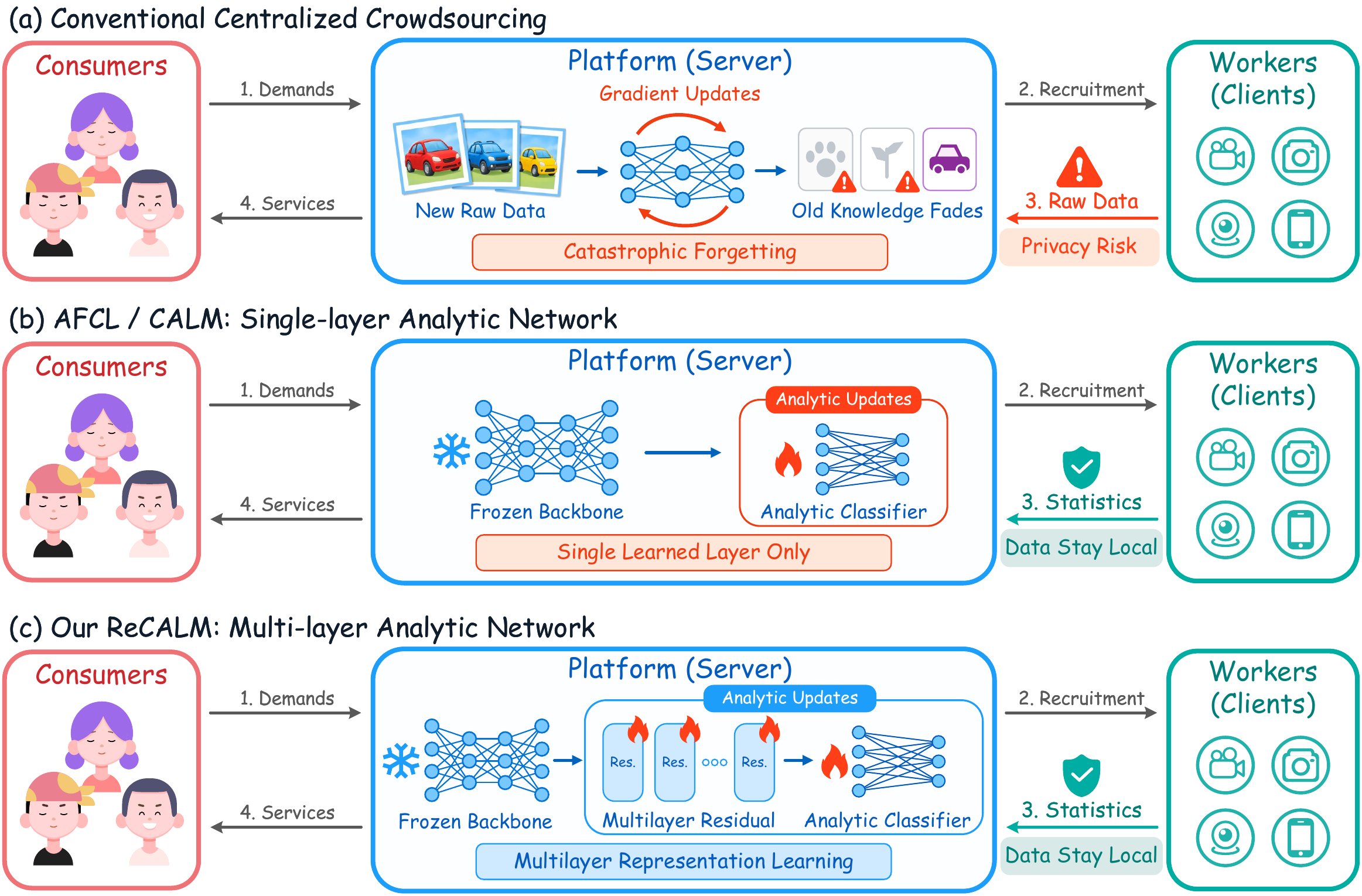}
    \vspace{-0.2cm}
    \caption{Comparison of crowdsourced service construction paradigms.
    (a) Conventional centralized crowdsourcing requires raw-data uploads.
    (b) Existing analytic methods for distributed continual scenarios learn a single-layer analytic classifier over fixed features.
    (c) ReCALM enables multi-layer analytic learning for distributed continual scenarios w/o uploading raw data.}
    \label{fig0:service-comparison}
    \vspace{-0.4cm}
\end{figure}

\begin{figure}[t]
    \centering
    \includegraphics[width=0.90\linewidth]{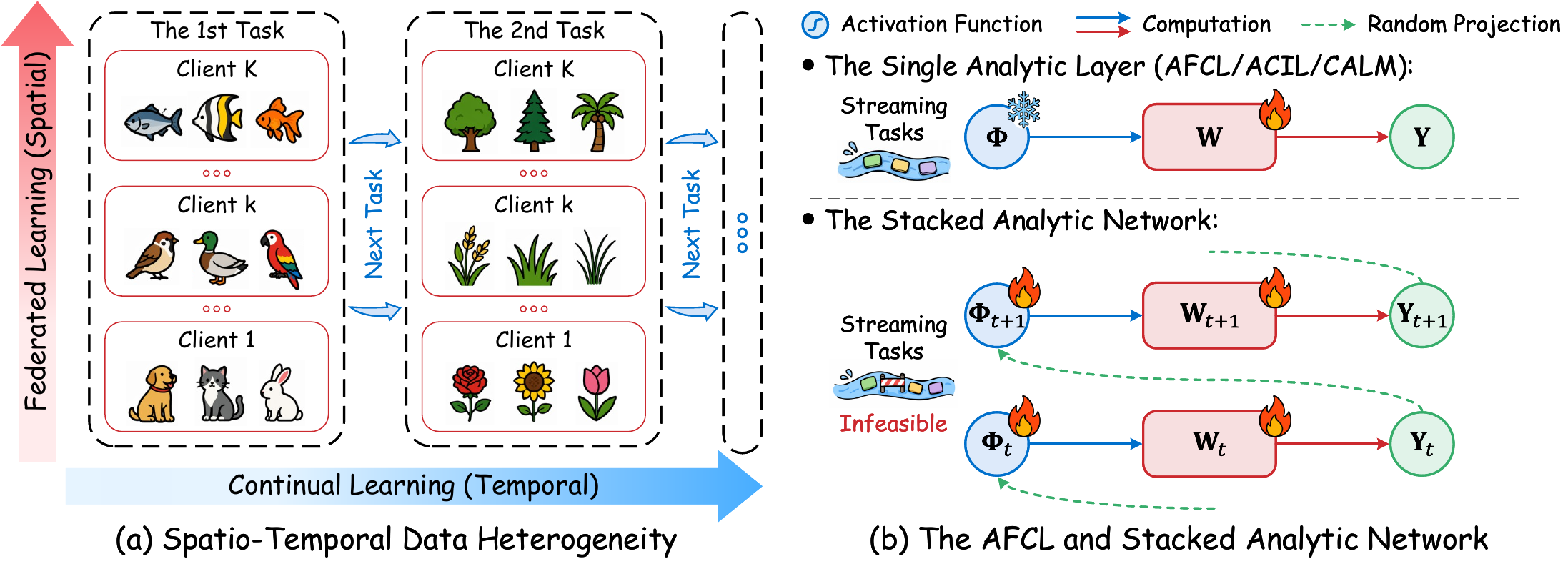}
    \vspace{-0.25cm}
    \caption{The distributed-continual-learning formulation and conventional analytic constructions. Panel (a) depicts data arriving from different participants and tasks, corresponding to clients and phases in our setting. Panel (b) shows a single-layer analytic classifier and the historical-feature dependence introduced by a conventional nonlinear stack.}
    \label{fig:fig1}
    \vspace{-0.4cm}
\end{figure}

Traditional approaches typically require workers to upload their private raw data for centralized training, incurring substantial communication overhead and privacy concerns~\cite{tang2026qdbpp,huang2025qlpdcs,tang2025cpdz}, as shown in Fig.~\ref{fig0:service-comparison}(a).
In contrast, distributed approaches, exemplified by federated learning, offer a promising alternative by leveraging locally computed statistics or model updates instead of directly transmitting raw data~\cite{fedavg,FL_2,AFL}.
In this paper, we focus on distributed continual learning, also referred to as Federated Continual Learning (FCL), where service models are incrementally updated without requiring workers to upload their raw data.
Despite its practical significance, FCL introduces a critical challenge, named Spatio-Temporal Data Heterogeneity (ST-DH)~\cite{FCL-0,FCL-3}.
This data heterogeneity (i.e., Non-IID data) manifests both \textit{spatially} (across different clients) and \textit{temporally} (across online tasks within the same client), resulting in a complex spatio-temporal interplay, as shown in Fig.~\ref{fig:fig1}(a).

As a direct adverse consequence of ST-DH, existing approaches often suffer from severe performance degradation~\cite{FCL-1,FCL-3,AFCL}.
Some FCL studies borrow terminology from continual learning and introduce the concept of catastrophic forgetting~\cite{FCL-1, FCL-3, AFCL,forgetting-FCL-1,forgetting-FCL-2}, which refers to the phenomenon that models lose previously learned knowledge when trained on new data~\cite{CIL_3_replay, ACIL_2, ACIL_new-2}. 
Specifically, this issue manifests not only as the temporal forgetting of previously learned tasks but also as the spatial erasure of worker-specific knowledge during global aggregation~\cite{FCL-1,FCL-3,AFCL}.
We call it spatio-temporal catastrophic forgetting, as detailed in the Appendix.

Recent studies have increasingly recognized that a key reason for catastrophic forgetting under ST-DH lies in the reliance on gradient-based updates, which may overwrite parameters encoding previously learned task-specific and worker-specific knowledge~\cite{FCL-2026,FCL-0,AFCL}.
Therefore, avoiding iterative gradient-based updates offers a promising direction to resolve the ST-DH challenge.
In this regard, analytic learning provides a gradient-free alternative that achieves heterogeneity invariance through recursive closed-form solutions, typically based on least-squares optimization~\cite{ACIL_L3A,ACIL_4}.
Along this line, AFL~\cite{AFL} was first proposed for distributed settings, followed by CALM~\cite{CALM} and AFCL~\cite{AFCL}, which extended this paradigm to distributed continual service construction.

Nevertheless, existing analytic methods are typically limited to single-layer linear networks.
Specifically, they construct a linear analytic classifier on features extracted by a frozen pre-trained backbone, which can be recursively updated to accommodate spatio-temporal incremental data.
Consequently, these methods can only capture a {\textit{linear mapping}} from the frozen features to the output, thus {\textit{fundamentally failing to perform deep representation learning}}.
On the other hand, the linear model also appears foundational to these analytic methods, as its simplicity with a convex optimization objective is a prerequisite for deriving its recursive analytical solution.
Our experiments in Section~\ref{sec:exp:rq2} also confirm that, even using a stronger backbone, the linear separability of its output features often remains insufficient and suboptimal in single-layer analytic networks.

This observation raises a pivotal research question: \textbf{\textit{Can we extend existing single-layer analytic networks into multi-layer networks that learn nonlinear representations layer by layer, while preserving recursive closed-form solutions and invariance to ST-DH?}}
Addressing this question involves two key technical challenges:
(1) constructing deep nonlinear analytic networks while maintaining closed-form solutions, and
(2) enabling distributed continual training under \textit{spatio-temporal data isolation} in continual service construction.
Together, these challenges make such an extension particularly nontrivial.

To illustrate the challenges more intuitively, let us consider a simple stacked analytic network as an example, as shown in Fig.~\ref{fig:fig1}(b).
This method uses the output of each analytic layer to construct the input to the next layer, with an activated random projection for introducing cross-layer non-linearity.
Yet, this simple idea is fundamentally infeasible in distributed continual service construction due to the conflict between the irreversible dependencies of features and the unavailability of historical data.
When learning a new task, the weights $\{\mathbf{W}_{l}\}_{l=0}^{L}$ require targeted updates, which directly influence the features $\{\mathbf{\Phi}_{l+1}\}_{l=0}^{L}$, while the nonlinear activation $\sigma$ renders the influence irreversible.
Since $\mathbf{\Phi}_{l+1}$ cannot be recomputed due to the inaccessibility of historical data in distributed continual settings, this method fails to support continual updating.

For similar reasons, existing methods that attempt to deepen analytic layers (such as DeepAFL and label encoding) are also infeasible in the distributed continual setting~\cite{LABEL, INS, DeepAFL}.
In fact, once analytic layers are stacked, the cross-layer features generally satisfy a functional relationship of the form $\mathbf{\Phi}_{l+1}=h(\mathbf{\Phi}_{l};\sigma)$.
In existing methods, the nonlinear activation function $\sigma$ is always applied to $\mathbf{\Phi}
_{l}$ to mimic the architecture of a Multi-Layer Perceptron (MLP).
While this operation is reasonable for meaningful deep representations, it destroys the linear dependencies among cross-layer features, thus breaking the recursive analytical solutions that existing analytic methods rely on.
For clarity, we provide more detailed analyses in Section~\ref{Section-3.2}.

\clearpage

To address these challenges, we propose \textbf{ReCALM}, a \textbf{Re}sidual-enhanced \textbf{C}rowdsourced \textbf{A}nalytic \textbf{L}earning \textbf{M}echanism that achieves deep nonlinear invariance to ST-DH for distributed continual service construction.
Inspired by the success of ResNet in gradient-based networks~\cite{ResNet}, we develop a {residual analytic network} with gradient-free residual blocks learned through closed-form solutions.
Specifically, we model layer-wise representation learning as $\mathbf{\Phi}_{l+1}=\mathbf{\Phi}_l+g_{l+1}$, where the nonlinear residual block $g_{l+1}$ is designed to be independent of the evolution of $\mathbf{\Phi}_l$.
Unlike conventional gradient-based networks that learn residual blocks through backpropagation, our residual blocks are learned analytically, preserving closed-form solutions while exhibiting three key properties: \textbf{\textit{nonlinearity}}, \textbf{\textit{stochasticity}}, and \textbf{\textit{learnability}}.
To implement distributed continual training under \textit{spatio-temporal data isolation}, we further devise a {cross-recursive strategy} that incorporates both layer-wise and phase-wise recursion via least squares, enabling continual analytic updates without accessing raw data.

\begin{samepage}
Our main contributions are summarized as follows:
\begin{enumerate}[itemsep=2pt,topsep=0pt,parsep=0pt]
    \item \textbf{Conceptually}, {drawing inspiration from residual connections, we propose ReCALM to achieve deep nonlinear invariance to ST-DH via multiple analytic layers for distributed continual service construction.}

    \item \textbf{Technically}, {we develop the gradient-free residual blocks for our ReCALM, enabling it to elegantly learn deep representations. Moreover, to implement distributed continual training, we propose a cross-recursive strategy that combines phase-wise statistical accumulation with layer-wise reconstruction.}

    \item \textbf{Theoretically}, 
    {we establish two properties of ReCALM: \textit{heterogeneity invariance} and \textit{deep representation}. To the best of our knowledge, it is the first time that these two desired properties have been observed to be simultaneously achieved in distributed continual service construction.}

    \item \textbf{Experimentally}, {we show ReCALM's superiority in accuracy and efficiency on three benchmarks.
    In particular, with 20 layers, ReCALM improves accuracy by 8.85\%–9.83\% over the single-layer analytic baseline, while reducing computational costs by over 94\% compared with gradient-based baselines.}
\end{enumerate}
\end{samepage}

\section{Related Work}
\label{sec:related-work}

\subsection{Crowdsourced Service Construction}
Mobile crowdsensing and computing combines observations from participants' devices to construct services across diverse environments~\cite{guo2015mobile,fan2025rmdfcv,tang2026eqtptd}.
The platform coordinates contributions according to service demands, while workers supply observations from their local contexts~\cite{wang2021csmc,zhao2024crowdgeo,fan2026pcqsp}.
CALM~\cite{CALM} develops this service-construction perspective for continually expanding recognition demands while protecting workers' data privacy.
It assigns local feature extraction and matrix computations to workers, while the platform accumulates their Gram-type statistics and recursively updates an analytic classifier without accessing raw sensing data.
Historical contributions are preserved in these accumulated statistics, enabling the resulting classifier to achieve equivalence to the centralized solution.
Nevertheless, CALM~\cite{CALM} learns only a single analytic classifier over fixed features, limiting its ability to refine feature representations.
In this paper, ReCALM extends the service model to a multi-layer residual analytic network and enables continual deep representation learning through its cross-recursive mechanism.

\subsection{Distributed Continual Learning}

As a popular distributed learning paradigm, federated learning has made significant progress~\cite{FL-hu1, fl-new-4, fl-new-5}.
By introducing the concept of continual learning, FCL has emerged as a pivotal topic for distributed dynamic environments, giving rise to a substantial body of influential studies~\cite{FCL-yichen-2, FCL-2026, FCL-yichen-3, FCL-1, FCL-new1}.
One of the most distinctive characteristics of FCL lies in spatio-temporal data isolation~\cite{FCL-yichen-2, FCL-2026, FCL-yichen-3, FCL-new1}.
Recently, increasing attention has been paid to the interaction within ST-DH in FCL, and the resulting degradation issue has been widely recognized as a critical challenge~\cite{FCL-0, FCL-1, FCL-4}.
One of the core causes of the ST-DH issue is widely recognized to lie in the reliance on gradient-based updates, which tend to overwrite well-learned model parameters~\cite{FCL-2026,FCL-0,AFCL}.
From this perspective, mitigating or avoiding excessive modification induced by gradients has become a popular direction, and analytic learning provides a promising technical tool~\cite{AFCL,CALM}.
However, existing analytic learning methods are significantly limited by their single-layer analytic models, which yield only linear invariance and suffer from underfitting.
Our ReCALM aims to address this issue by deepening the analytic networks.

\subsection{Analytic Learning}

Analytic learning computes model parameters through closed-form solutions, with least-squares estimation providing a common basis~\cite{AL_new_0,ACIL_1,acu}.
Conventional gradient-based distributed learning faces several challenges.
Specifically, \textbf{(1) Heterogeneity:} Non-IID data across workers may lead to conflicting local gradient updates, degrading model performance~\cite{FedProx,FL_2}.
\textbf{(2) Scalability:} Increasing numbers of participating workers make iterative model aggregation and coordination more challenging~\cite{FL_2}.
\textbf{(3) Convergence:} Gradient-based optimization may converge slowly, particularly under heterogeneous data and limited communication rounds~\cite{FedProx}.
\textbf{(4) Efficiency:} Repeated local training and multi-round aggregation incur substantial computation and communication overhead~\cite{fedavg,AFL}.
These challenges are particularly pronounced in crowdsourced continual service construction, where heterogeneous worker data and evolving service demands may exacerbate spatio-temporal catastrophic forgetting during gradient-based model updates~\cite{FCL-3,AFCL}.
By replacing iterative gradient updates with closed-form computations, analytic learning offers a promising alternative to address these challenges~\cite{AFL,AFCL,fan2026anapfl}.

\begin{table*}[t]
\centering
\caption{Comparison of representative paradigms and
analytic learning methods for continual service construction.}
\label{tab:service-capabilities}
\vspace{-0.1cm}

\footnotesize
\setlength{\tabcolsep}{4pt}
\setlength{\lightrulewidth}{0.35pt}
\setlength{\cmidrulewidth}{0.35pt}
\renewcommand{\arraystretch}{1.2}
\renewcommand{\tabularxcolumn}[1]{m{#1}}

\begin{tabularx}{\textwidth}{
@{}
>{\raggedright\arraybackslash}m{0.15\textwidth}
*{4}{>{\centering\arraybackslash}X}
>{\columncolor{myblue!5}\centering\arraybackslash}X
@{}
}
\toprule

\multirow{2}{*}{\textbf{Property}}
& \multicolumn{2}{c}{\textbf{Conventional Paradigms}}
& \multicolumn{3}{c}{\textbf{Analytic Learning Methods}} \\

\cmidrule(lr){2-3}
\cmidrule(lr){4-6}

&
\textbf{Centralized}
& \makecell{\textbf{Distributed}}
& \makecell{\textbf{DeepAFL}~\cite{DeepAFL}}
& \makecell{\textbf{CALM}~\cite{CALM}}
& \textbf{ReCALM} \\

\midrule

\textbf{Worker Uploads}
& \makecell{Raw data}
& \makecell{Weights or gradients}
& Local statistics
& Local statistics
& Local statistics \\

\textbf{Worker-side Computation}
& Data sensing
& \makecell{Heavy BP training}
& \makecell{Lightweight\\matrix operations}
& \makecell{Lightweight\\matrix operations}
& \makecell{Lightweight\\matrix operations} \\

\textbf{Platform-side Computation}
& \makecell{Heavy BP training}
& \makecell{Iterative model\\aggregation}
& \makecell{Lightweight\\matrix updates}
& \makecell{Lightweight\\matrix updates}
& \makecell{Lightweight\\matrix updates} \\

\midrule

\textbf{Continual Service Construction}
& \makecell{Via CL\\extensions}
& \makecell{Via FCL\\extensions}
& Not established
& Native
& {Native} \\

\textbf{Continual Service Quality}
& \makecell{Catastrophic\\forgetting}
& \makecell{Catastrophic\\forgetting}
& Not established
& \makecell{Exact analytic\\retention}
& \makecell{{Exact analytic}\\{retention}} \\

\midrule

\textbf{Analytic Data Setting}
& ---
& ---
& \makecell{Static data}
& \makecell{Continual data}
& \makecell{Continual data} \\

\textbf{Trainable Components}
& \makecell{multi-layer\\network}
& \makecell{multi-layer\\network}
& \makecell{multi-layer\\network}
& \makecell{Single\\analytic head}
& \makecell{multi-layer\\network} \\

\bottomrule
\end{tabularx}

\end{table*}

Recently, CALM~\cite{CALM} and AFCL~\cite{AFCL} have applied recursive analytic learning to distributed continual scenarios.
Since both methods can ultimately be reduced to a single-layer analytic classifier with fixed features, they can be considered functionally equivalent.
DeepAFL~\cite{DeepAFL} further extends analytic learning to multi-layer residual networks and establishes centralized equivalence for fixed distributed datasets.
Notably, DeepAFL~\cite{DeepAFL} fails to work in continual learning scenarios because updating preceding layers changes historical representations required by subsequent layers, as further detailed in the Appendix.
To overcome these limitations, ReCALM introduces a novel residual analytic architecture and a cross-recursive strategy, enabling continual multi-layer analytic learning while preserving equivalence to centralized analytic learning.
Interestingly, beyond enabling continual deep learning, ReCALM achieves better performance than DeepAFL~\cite{DeepAFL}, as detailed in the Appendix.
For clarity, Table~\ref{tab:service-capabilities} provides detailed comparisons between our ReCALM and representative methods.

\section{System Model and Problem Formulation}
\label{sec:3.1}

We consider a typical crowdsourced continual service construction scenario involving three entities: a \textit{crowdsourcing platform} (i.e., the server), a large number of \textit{sensing workers} (i.e., the clients), and \textit{service consumers}.
Consumers continuously introduce new or evolving service demands, requiring the server to progressively extend its existing services while preserving previously acquired capabilities.
To accommodate these incremental demands, the server continually recruits sensing workers to collect new observations and contribute data for updating the service model.
Through this recurring process of demand arrival, client participation, and model updating, the server achieves crowdsourced continual service construction to satisfy consumers' evolving requirements.

Formally, we consider $K$ sensing workers and divide the continual service construction process into $T$ phases, indexed by $t \in \{1,2,\cdots,T\}$.
For each phase $t$, let $S_t$ denote the set of participating clients, which may vary across service phases.
For clarity, we consider an image-recognition service as a representative example, while our mechanism can be extended to other service tasks.
For each client $k \in S_t$, its local dataset is denoted as $\mathcal{D}_{t,k}=\{\mathbf{X}_{t,k},\mathbf{Y}_{t,k}\}$, where $\mathbf{X}_{t,k}\in\mathbb{R}^{N_{t,k}\times d_\text{H}\times d_\text{W}\times d_\text{C}}$ and $\mathbf{Y}_{t,k}\in\mathbb{R}^{N_{t,k}\times C_t}$ represent the locally collected images and their corresponding labels, respectively.
$N_{t,k}$ denotes the number of local samples, while $d_\text{H}$, $d_\text{W}$, and $d_\text{C}$ represent the image height, width, and number of channels, respectively.
Meanwhile, $C_t$ denotes the cumulative number of recognition classes introduced up to phase $t$, reflecting the evolving service demands.
The complete dataset is defined as $\mathcal{D}=\bigcup_{t=1}^{T}\bigcup_{k\in S_t}\mathcal{D}_{t,k}$, where $\mathcal{D}_{t,k}\cap\mathcal{D}_{t',k'}=\emptyset$ for all $(t,k)\neq(t',k')$.
Our objective is to construct an $L$-layer deep residual analytic network as the service model, enabling richer feature representations while preserving equivalence to centralized analytic learning over all accumulated data.
We use the terms clients and server following the distributed-learning terminology used by related methods.
For clarity, the primary notations are summarized in the Appendix.

\section{Residual Analytic Network in Centralized Static Scenario}
\label{sec:recalm-learning}

In this section, we introduce the residual analytic network of our ReCALM from a centralized static perspective.
Specifically, we consider the complete dataset $\mathcal{D}=\{\mathbf{X},\mathbf{Y}\}$ and illustrate how to progressively construct the $L$-layer residual analytic network through layer-wise closed-form computations.
Following AFCL~\cite{AFCL} and CALM~\cite{CALM}, we employ a frozen pre-trained backbone for initial feature extraction.
This shared feature mapping provides a consistent representation space for subsequent network construction and continual updates.
Specifically, we first extract the initial features $\Phi_0$ from the raw input $\mathbf{X}$ as follows:
\begin{equation}
\label{eq:section3.1-1}
    \mathbf{\Phi}_{0} = \sigma(\text{Backbone}(\mathbf{X}, \mathbf{\Theta}) \mathbf{A}),
\end{equation}
where $\text{Backbone}(\cdot, \mathbf{\Theta})$ is the selected pre-trained backbone with parameters $\mathbf{\Theta}$.
Meanwhile, $\sigma(\cdot)$ denotes the nonlinear activation function, and $\mathbf{A} \in \mathbb{R}^{d_\text B \times d_\Phi}$ is a random projection matrix. 
The employment of $\sigma(\cdot)$ and $\mathbf{A}$ projects the features extracted by the backbone into a high-dimensional nonlinear space, thereby preliminarily enhancing their linear separability~\cite{ACIL_6, MiN, ACIL_new-1}.

Subsequently, we proceed to construct the $L$-layer analytic network for deep representation learning.
ReCALM adopts a layer-wise forward construction process, which recursively determines the classifier $\mathbf{W}_l$ using this layer's features $\mathbf{\Phi}_l$, and subsequently constructs the next layer's features $\mathbf{\Phi}_{l+1}$ to fit the residuals derived from the current layer.
For clarity, we focus on the construction of an arbitrary intermediate layer $l \in [0, L)$, assuming that all preceding layers and features $\mathbf{\Phi}_l$ have been constructed via a similar recursive procedure.

Specifically, we first construct the analytic classifier $\mathbf{W}_l$ for the layer $l$.
Consistent with existing analytic learning methods, the optimization objective for $\mathbf{W}_l$ can be formulated as follows:
\begin{equation}
\label{eq:section3.1-2}
  \mathbf{W}_l = \arg\underset{\mathbf{W}}{\min} \ 
  \|\mathbf{Y}-\mathbf{\Phi}_{l}\mathbf{W}\|^2_\text{F}+\mathbf{\lambda} \| \mathbf{W} \|^{2}_\text{F},
\end{equation}
where $\| \cdot \|^{2}_\text{F}$ denotes the Frobenius norm and $\lambda > 0$ represents the regularization parameter.
We employ the Mean Squared Error (MSE) loss here because it admits a closed-form least-squares solution.
Indeed, MSE is widely adopted in existing analytic learning methods and has shown comparable performance to cross-entropy loss~\cite{MSE-0, MSE-1}, with improved robustness to non-IID data~\cite{MiN, ACIL_new-1,DeepAFL}.
By applying the least squares method, we can directly derive the optimal analytical solution \eqref{eq:section3.1-3} to the objective \eqref{eq:section3.1-2}, and the detailed proof is provided in \textbf{Lemma 1} of the Appendix.
Thus, the predictor $\mathbf{W}_l$ can be computed directly via \eqref{eq:section3.1-3}.
\begin{equation}
\label{eq:section3.1-3}
  \mathbf{W}_l = (\mathbf{\Phi}_l^\top \mathbf{\Phi}_l + \lambda\mathbf{I})^{-1} \mathbf{\Phi}_l^\top \mathbf{Y}.
\end{equation}

Then, based on the optimized $\mathbf{W}_l$, we proceed to update the feature representations to obtain $\mathbf{\Phi}_{l+1}$.
Inspired by ResNet~\cite{ResNet}, we employ similar skip connections to refine the features layer-by-layer as follows:
\begin{equation}
\label{eq:section3.1-4}
    \mathbf{\Phi}_{l+1} =  \mathbf{\Phi}_{l} + g_{l+1}(\mathbf{\Phi}_{0}), \forall l \in [0,L),
\end{equation}
where $g_{l+1}(\cdot)$ denotes the residual block, and the specific design is detailed in subsequent~\eqref{eq:section3.1-5}.
Notably, we deliberately anchor the input of each residual block to the fixed initial features $\mathbf{\Phi}_{0}$.
This design is essential to the feasibility of the subsequent cross-recursive strategy for crowdsourced continual service construction.
Meanwhile, the skip connection carries the learned features $\mathbf{\Phi}_{l}$ between adjacent layers, thereby enabling deeper layers to refine the representations.

To support deep representation learning, we design the residual blocks to exhibit properties of \textit{\textbf{nonlinearity}}, \textit{\textbf{stochasticity}}, and \textit{\textbf{learnability}}.
Thus, in ReCALM, the specific residual block $g_{l+1}( \cdot )$ is designed as follows:
\begin{equation}
\label{eq:section3.1-5}
    g_{l+1}(\mathbf{\Phi}_{0}) =  \sigma(\mathbf{\Phi}_{0} \mathbf{B}_l) \mathbf{\Omega}_{l+1} = \mathbf{F}_{l}\mathbf{\Omega}_{l+1},
\end{equation}
where $\sigma(\cdot)$ denotes the nonlinear activation function, which serves as a critical component in deep representation learning by introducing \textit{\textbf{nonlinearity}}.
The term $\mathbf{B}_l \in \mathbb{R}^{d_\Phi \times d_\text{F}}$ is the random projection matrix that introduces \textit{\textbf{stochasticity}}, analogous to the randomness inherent in SGD for gradient-based methods.
Furthermore, $\mathbf{\Omega}_{l+1} \in \mathbb{R}^{d_\text{F} \times d_\Phi}$ signifies the learnable transformation matrix responsible for \textit{\textbf{learnability}}.
For brevity, we denote the hidden feature matrix of layer $l$ as $\mathbf{F}_{l} = \sigma(\mathbf{\Phi}_{0} \mathbf{B}_l)$.
In our experiments, we provide extensive ablation studies to show the effectiveness of these components.

To obtain the optimal feature representation $\mathbf{\Phi}_{l+1}$, we proceed to determine the optimal learnable transformation matrix $\mathbf{\Omega}_{l+1}$.
Specifically, analogous to~\eqref{eq:section3.1-2}, we employ the MSE loss to perform empirical risk minimization for the updated features $\mathbf{\Phi}_{l+1} = \mathbf{\Phi}_{l} + \mathbf{F}_{l}\mathbf{\Omega}_{l+1}$, as follows.
\begin{equation}
    \begin{aligned}
\label{eq:section3.1-6}
\mathbf{\Omega}_{l+1} &= \arg\underset{\mathbf{\Omega}}{\min} \ \|\mathbf{Y}-(\mathbf{\Phi}_{l} + \mathbf{F}_{l}\mathbf{\Omega})\mathbf{W}_l\|^2_\text{F}+\mathbf{\gamma} \| \mathbf{\Omega} \|^{2}_\text{F}  = \arg\underset{\mathbf{\Omega}}{\min} \ \|\mathbf{R}_l-\mathbf{F}_{l}\mathbf{\Omega}\mathbf{W}_l\|^2_\text{F}+\mathbf{\gamma} \| \mathbf{\Omega} \|^{2}_\text{F}.
\end{aligned}
\end{equation}

Here, we denote $\mathbf{R}_l = \mathbf{Y}-\mathbf{\Phi}_{l}\mathbf{W}_l$ as the residual matrix for the $l$-th layer to simplify the notation.
Based on this, the optimization objective~\eqref{eq:section3.1-6} can be viewed as a special case of generalized Sylvester matrix equations~\citep{sand-1, sand-2, sand-3, feat-boost}, where the unknown variable $\mathbf{\Omega}$ is \textit{sandwiched} between two known ones, $\mathbf{F}_l$ and $\mathbf{W}_l$.
Thus, we term it as the \textit{sandwiched least squares problem}.
Using the least squares, we can obtain its analytical solution in~\eqref{eq:section3.1-7}, with the proof provided in \textbf{Lemma 2} of the Appendix.
\begin{equation}
\label{eq:section3.1-7}
    \mathbf{\Omega}_{l+1}
    =\mathbf{V}_l[(\mathbf{V}_l^\top \mathbf{F}_l^\top \mathbf{R}_l \mathbf{W}_l^\top \mathbf{U}_l) \oslash \mathbf{T}_l ] \mathbf{U}_l^\top, \quad \mathbf{T}_l = \gamma\mathbf{1} + \text{diag}(\mathbf{\Lambda}_l^\text{H}) \otimes \text{diag}(\mathbf{\Lambda}_l^\text{W}),
\end{equation}
where $\mathbf{F}_{l}^\top\mathbf{F}_{l} = \mathbf{V}_l\mathbf{\Lambda}_l^\text{H}\mathbf{V}_l^\top$ and $\mathbf{W}_{l}\mathbf{W}_{l}^\top = \mathbf{U}_l\mathbf{\Lambda}_l^\text{W}\mathbf{U}_l^\top$ represent the spectral decompositions.
The operators $\oslash$ and $\otimes$ denote element-wise division and the outer product, respectively.
Upon deriving the optimal learnable transformation matrix $\mathbf{\Omega}_{l+1}$, we can subsequently obtain the updated feature matrix $\mathbf{\Phi}_{l+1}$ for the next layer by $\mathbf{\Phi}_{l+1} =  \mathbf{\Phi}_{l} + \mathbf{F}_{l}\mathbf{\Omega}_{l+1}$, according to \eqref{eq:section3.1-4} and \eqref{eq:section3.1-5}.

The aforementioned process is executed layer-by-layer until the complete $L$-layer network is constructed, following the sequence $\mathbf{W}_0 \to \mathbf{\Omega}_1 \to \mathbf{W}_1 \to \cdots \to \mathbf{\Omega}_L \to \mathbf{W}_L$.
Building on this centralized construction, we will then introduce the distributed continual training strategy of our ReCALM in subsequent Section~\ref{Section-5}.

\section{Cross-Recursive Strategy for Distributed Continual Training}
\label{Section-5}

Specifically, we first introduce the core idea of our proposed cross-recursive strategy in Section~\ref{Section-3.2}.
We then derive the eight Gram-type statistics required for its implementation in Section~\ref{sec:implementation-statistics}, followed by the detailed procedures on the client side and server side in Sections~\ref{sec:implementation-worker} and~\ref{sec:implementation-platform}, respectively.
Furthermore, in Section~\ref{sec:worker-participation}, we describe how ReCALM accommodates dynamic client participation and asynchronous updates.

\subsection{Basic Idea}
\label{Section-3.2}

Here, we first introduce how our ReCALM can be extended to the distributed continual setting, where data are distributed across clients, and new contributions arrive over successive service phases.
Specifically, we elaborate on the key idea of our cross-recursive strategy, which enables the continual construction of the multi-layer analytic network from accumulated client statistics.
To facilitate the following presentation, we first clarify the notations used for service phases, network layers, and clients.
The superscript $t \in [1,T]$ denotes the service-phase index for model weights, while the subscript $l \in [0,L]$ denotes the layer index of ReCALM.
Thus, we denote the analytic classifier and the transformation matrix in phase $t$ at layer $l$ by $\mathbf{W}^t_l$ and $\mathbf{\Omega}^t_l$, respectively.
Moreover, the superscript $k \in S_t$ is used to indicate local quantities associated with client $k$ at phase $t$.

For intermediate features $\mathbf{\Phi}$, in addition to the \textbf{\textit{superscript}} $t\in [1,T]$, which indicates that the features are computed using the weights (i.e., $\{\mathbf{W}_{l}^{t}\}$ and $\{\mathbf{\Omega}_{l}^{t}\}$) updated up to phase $t$, we also introduce another phase index $t'\in [1,T]$ in the \textbf{\textit{subscript}} to indicate the phase from which the data are drawn.
Note that when describing the features at a specific phase, we usually have $t=t'$. In this case, the subscript $t'$ can be omitted for brevity when no ambiguity arises. 
For instance, $\mathbf{\Phi}^{t,k}_l=\mathbf{\Phi}^{t,k}_{l,t}$ denotes the local feature matrix of client $k$ at layer $l$, using the data and weights both in phase $t$.

It is worth noting that when the superscript $k$ is omitted, the variable denotes a global aggregated quantity obtained by vertically concatenating all clients' local quantities.
Meanwhile, symbols with a \textit{tilde} denote the corresponding global quantities accumulated by all historical data and participating clients up to phase $t$, using the newest model weights.
For instance, the global aggregated feature ${\mathbf{\Phi}}_{l}^t$ and the global accumulated feature $\widetilde{\mathbf{\Phi}}_{l}^t$ can be formulated as follows:
\begin{equation}
\label{eq:section3.2-5}
\mathbf{\Phi}_{l}^t= 
\mathbf{\Phi}_{l,t}^t=
\text{Concat}[(\mathbf{\Phi}_{l,t}^{t,k})^\top]_{k \in S_t}^\top,
 \quad
\mathbf{\widetilde \Phi}_{l}^t = 
\mathbf{\Phi}_{l,(1:t)}^{t}=
\text{Concat}[(\mathbf{\Phi}_{l,t'}^{t})^\top]_{1 \leq t' \leq t}^\top.
\end{equation}

As discussed earlier, one major technical challenge in extending analytic learning to multi-layer networks for crowdsourced continual service construction is that the server cannot directly access clients' raw data, nor can it require clients to retain historical data.
Specifically, when new client contributions arrive, the multi-layer network must be updated using newly received statistics and previously accumulated information.
However, layer-wise updates also affect the historical features required for phase-wise accumulation.
For example, updating the weights of preceding layers changes the intermediate representations of historical samples, making previously accumulated features outdated.
These features cannot be recomputed under the updated weights as historical data are no longer accessible.
Thus, for $l \ge 1$, the straightforward strategy of concatenating previously accumulated features with current-phase features fails, as it cannot recover the correct updated global features:
\begin{equation}
\label{eq:section3.2-neq}
\text{Concat}[ (\mathbf{\widetilde \Phi}_{l}^{\myredA{t-1}})^\top , (\mathbf{\Phi}_{l}^{\myblueA{t}})^\top]^\top
= \text{Concat}[ (\mathbf{\Phi}_{l,1}^{\myredA{t-1}})^\top,
(\mathbf{\Phi}_{l,2}^{\myredA{t-1}})^\top,
\cdots,
(\mathbf{\Phi}_{l,t-1}^{\myredA{t-1}})^\top,
(\mathbf{\Phi}_{l,t}^{\myblueA{t}})^\top]^\top
\neq
\mathbf{\widetilde \Phi}_{l}^{\myblueA{t}}.
\end{equation}

To address the preceding challenges, let's revisit our specific closed-form expressions.
Our core goal is to compute and update the two sets of weights (i.e., $\{\mathbf{W}_{l}^{t}\}$ and $\{\mathbf{\Omega}_{l}^{t}\}$) based on \eqref{eq:section3.1-3} and \eqref{eq:section3.1-7}, where the computation depends on certain data-related features or labels (such as $\widetilde{\mathbf{\Phi}}_l^{t}$, $\widetilde{\mathbf{F}}_l^{t}$, $\widetilde{\mathbf{Y}}^{t}$, and $\widetilde{\mathbf{R}}_l^{t}$).
Our key insight here is that the required dependence on features and labels can be captured by certain inner-product statistics (also called Gram-type statistics), e.g., $(\widetilde{\mathbf{\Phi}}_l^{t})^\top
\widetilde{\mathbf{\Phi}}_l^t$, which suffice for the overall computation.
For example, we can define several basic Gram-type statistics as follows.
\begin{equation}
          \myredA{\mathbf{\widetilde G}_{l}^{t}} = (\mathbf{\widetilde \Phi}_{l}^{t})^{\top}\mathbf{\widetilde \Phi}_{l}^{t},
          \quad
          \myredA{\mathbf{\widetilde H}_{l}^{t}} = (\mathbf{\widetilde \Phi}_{l}^{t})^{\top}\mathbf{\widetilde Y}^{t}, 
          \quad
          \myredA{\mathbf{\widetilde N}_{l}^{t}}= (\mathbf{\widetilde F}_{l}^{t})^{\top}\mathbf{\widetilde F}_{l}^{t},
          \quad
          \myredA{\widetilde{\bm{\Upsilon}}_{l}^{t}}= (\mathbf{\widetilde F}_{l}^{t})^{\top}\mathbf{\widetilde R}_{l}^{t}.
\end{equation}

Then, based on \eqref{eq:section3.1-3} and \eqref{eq:section3.1-7}, we can reformulate the weights computation processes for the analytic classifier $\mathbf{W}_l^t$ and the transformation matrix $\mathbf{\Omega}_{l+1}^t$ as follows.
\begin{equation}
     \mathbf{W}_l^t  = 
     (\myredA{\mathbf{\widetilde G}_l^t}  + \lambda\mathbf{I})^{-1} \myredA{\mathbf{\widetilde H}_l^t},
\quad 
     \mathbf{\Omega}_{l+1}^t
    =\mathbf{V}_{l}^t[(\mathbf{V}_{l}^{t \top} \myredA{\widetilde{\bm{\Upsilon}}_{l}^t} \mathbf{W}_{l}^{t \top} \mathbf{U}_{l}^t) \oslash \mathbf{T}_{l}^t ] \mathbf{U}_{l}^{t \top}.
\end{equation}
Here, $\mathbf{T}_{l}^{t} = \gamma\mathbf{1} + \text{diag}(\mathbf{\Lambda}_{l,\text{H}}^{t}) \otimes \text{diag}(\mathbf{\Lambda}_{l,\text{W}}^{t})$ is a change-of-variable term, while $\myredA{\mathbf{\widetilde N}_{l}^{t}} = \mathbf{V}_{l}^{t}\mathbf{\Lambda}_{l, \text H}^{t}\mathbf{V}_{l}^{t \top}$ and $\mathbf{W}_{l}^{t}\mathbf{W}_{l}^{t \top} = \mathbf{U}_{l}^{t}\mathbf{\Lambda}_{l, \text W}^{t}\mathbf{U}_{l}^{t \top}$ represent the spectral decompositions.
In this way, we can recover the model weights from Gram-type statistics without accessing individual features or labels, and thus focus on computing and updating these statistics.
Preserving Gram-type statistics also underlies the recursive analytic classifiers in CALM~\cite{CALM}.
The privacy of Gram-type statistics can be further enhanced by applying secure aggregation protocols when aggregating local statistics, as also shown in~\cite{DeepAFL,Fed3R}.

\begin{figure}[t]
    \centering
    \includegraphics[width=1.0\linewidth]{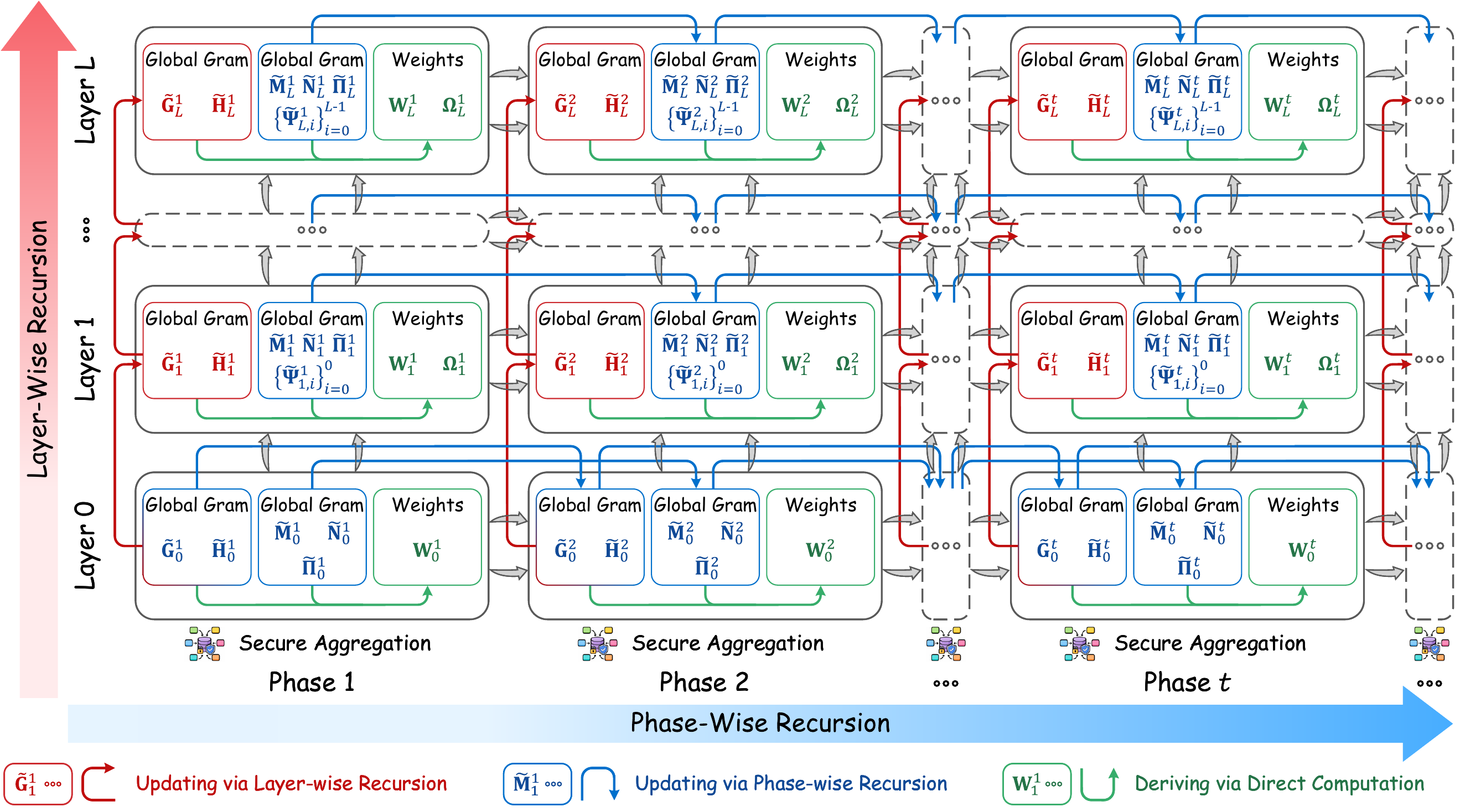}
    \caption{The cross-recursive strategy of our ReCALM.}
    \vspace{-0.1cm}
    \label{fig:fig2}
\end{figure}

Next, the key technical challenge lies in how to recursively update the Gram-type statistics.
In ReCALM, both layer-wise and phase-wise recursion are involved, and we develop a cross-recursive strategy that combines them, as illustrated in Fig.~\ref{fig:fig2}.
The complete update procedure involves the recursion of six core Gram-type statistics and the computation of two auxiliary intermediate statistics.
\textbf{\textit{Thus, we mainly present the key design rationale here, leaving the detailed derivations and computational procedures to subsequent Sections~\ref{sec:implementation-statistics} to \ref{sec:implementation-platform}.}}

In our cross-recursive strategy, the key criterion for determining whether a statistic should follow the layer-wise or phase-wise update path is whether it involves the deep features $\mathbf{\widetilde \Phi}_{l}^{t}$ with $l\ge1$.
When a statistic involves only $\mathbf{\widetilde \Phi}_{0}^{t}$, $\mathbf{\widetilde Y}^{t}$, $\mathbf{\widetilde F}_{l}^{t}$ that are independent of model weights, we can update the inner-product statistics via phase-wise summation.
Let's take $\mathbf{\widetilde G}_{0}^{t}$ as an example:
\begin{equation}
\label{eq:3.3-12-new}
\myredA{\mathbf{\widetilde G}_{0}^{t}} =
(\mathbf{\widetilde \Phi}_{0}^{t})^{\top}\mathbf{\widetilde \Phi}_{0}^{t} =
\myredA{\mathbf{\widetilde G}_{0}^{t-1}} + \sum\nolimits_{k \in S_t}\mathbf{G}_{0}^{t,k}.
\end{equation}

For statistics involving $\mathbf{\widetilde \Phi}_{l}^{t}$ with $l\ge1$, however, phase-wise concatenation leaves the historical features tied to earlier model weights, as revealed in \eqref{eq:section3.2-neq}.
Therefore, we recompute these statistics based on the feature relationships between adjacent layers under the current model weights.
We illustrate this using $\mathbf{\widetilde G}_{l}^{t}$ with $l\ge1$ as an example:
\begin{equation}
\label{eq:layer-wise-update-G}
\begin{aligned}
    \myredA{\widetilde{\mathbf{G}}_l^t} &= \widetilde{\mathbf{\Phi}}_{l}^{t\top}
    \widetilde{\mathbf{\Phi}}_{l}^{t}=
    (\widetilde{\mathbf{\Phi}}_{l-1}^{t}
    +
    \widetilde{\mathbf{F}}_{l-1}^{t}\mathbf{\Omega}_{l}^{t})^{\top}
    (\widetilde{\mathbf{\Phi}}_{l-1}^{t}
    +
    \widetilde{\mathbf{F}}_{l-1}^{t}\mathbf{\Omega}_{l}^{t}) \\
    &=
    (\widetilde{\mathbf{\Phi}}_{l-1}^{t})^{\top}
    \widetilde{\mathbf{\Phi}}_{l-1}^{t}
    +
    (\widetilde{\mathbf{\Phi}}_{l-1}^{t})^{\top}
    \widetilde{\mathbf{F}}_{l-1}^{t}
    \mathbf{\Omega}_{l}^{t}
    +
    (\mathbf{\Omega}_{l}^{t})^{\top}
    (\widetilde{\mathbf{F}}_{l-1}^{t})^{\top}
    \widetilde{\mathbf{\Phi}}_{l-1}^{t}
    +
    (\mathbf{\Omega}_{l}^{t})^{\top}
    (\widetilde{\mathbf{F}}_{l-1}^{t})^{\top}
    \widetilde{\mathbf{F}}_{l-1}^{t}
    \mathbf{\Omega}_{l}^{t} \\
    &= 
    \myredA{\mathbf{\widetilde G}_{l-1}^{t}} + (\myredA{\mathbf{\widetilde{\Gamma}}_{l-1}^{t}})^{\top}\mathbf{\Omega}_{l}^{t} +
    (\mathbf{\Omega}_{l}^{t})^{\top}{\myredA{\widetilde{\mathbf\Gamma}_{l-1}^{t}}} +
    (\mathbf{\Omega}_{l}^{t})^{\top}{\myredA{\mathbf{\widetilde N}_{l-1}^{t}}}\mathbf{\Omega}_{l}^{t}. 
\end{aligned}
\end{equation}
Here, ${\myredA{\widetilde{\mathbf\Gamma}_{l-1}^{t}}}=(\widetilde{\mathbf{F}}_{l-1}^{t})^{\top} \widetilde{\mathbf{\Phi}}_{l-1}^{t}$ is a newly introduced Gram-type statistic.
As this formulation involves only three Gram-type statistics (i.e., $\myredA{\mathbf{\widetilde G}_{l-1}^{t}}$, $\myredA{\mathbf{\widetilde{\Gamma}}_{l-1}^{t}}$, and $\myredA{\mathbf{\widetilde N}_{l-1}^{t}}$) from layer $(l-1)$ to derive the next-layer Gram-type statistic $\myredA{\widetilde{\mathbf{G}}_l^t}$ at layer $l$, we refer to this procedure as a layer-wise update.
Moreover, we can observe that the known model weight $\mathbf{\Omega}_{l}^{t}$ appears only outside the Gram-type statistics, via left or right multiplication.
Thus, once all the required Gram-type statistics have been recursively computed, we can directly compute and update $\myredA{\widetilde{\mathbf{G}}_l^t}$ sequentially according to this layer-wise formulation.

Notably, based on \eqref{eq:3.3-12-new}--\eqref{eq:layer-wise-update-G}, the update form of $\myredA{\mathbf{\widetilde G}_{l}^{t}}$ differs between the two cases $l=0$ and $l\ge1$.
This is because the former corresponds to frozen features and is independent of the model weights, whereas the latter depends on the weights and thus undergoes more complex changes in each phase.
The decision to anchor the input of each residual block to $\mathbf{\Phi}_{0}$ in \eqref{eq:section3.1-4}--\eqref{eq:section3.1-5} reflects a co-design of the network architecture and training mechanism.
This design keeps the nonlinear feature maps fixed across service phases.
The learned features $\mathbf{\Phi}_{l}$ are updated through linear combinations of these fixed maps, allowing the Gram-type statistics to be reconstructed across layers.
Following the preceding insights, our cross-recursive strategy establishes the update relationships among eight Gram-type statistics, 
allowing our ReCALM to update its multi-layer service model continually while preserving the centralized analytic solution.
We will present the detailed implementation in the next section.

\subsection{Gram-Type Statistics for Detailed Implementation}
\label{sec:implementation-statistics}

Building on the cross-recursive strategy introduced in Section~\ref{Section-3.2}, we now specify the eight Gram-type statistics required for the distributed implementation of ReCALM.
These statistics summarize the data-dependent information needed to recursively update the multi-layer analytic network without accessing historical raw data.
As discussed in Section~\ref{Section-3.2}, four of these statistics have already been introduced to express the closed-form computations of the analytic classifiers and transformation matrices, as follows:
\begin{equation}
\label{eq:app-Implementation-1}
    \widetilde{\mathbf{H}}_{l}^{t}
    =
    (\widetilde{\mathbf{\Phi}}_{l}^{t})^{\top}
    \widetilde{\mathbf{Y}}^{t},
    \quad
    \widetilde{\mathbf{G}}_{l}^{t}
    =
    (\widetilde{\mathbf{\Phi}}_{l}^{t})^{\top}
    \widetilde{\mathbf{\Phi}}_{l}^{t},
    \quad
    \widetilde{\mathbf{N}}_{l}^{t}
    =
    (\widetilde{\mathbf{F}}_{l}^{t})^{\top}
    \widetilde{\mathbf{F}}_{l}^{t},
    \quad
    \widetilde{\bm{\Upsilon}}_{l}^{t}
    =
    (\widetilde{\mathbf{F}}_{l}^{t})^{\top}
    \widetilde{\mathbf{R}}_{l}^{t}.
\end{equation}

But, these four statistics alone are insufficient to support the complete recursive updates across service phases and network layers.
In particular, reconstructing the statistics associated with updated deep representations requires additional information about the relationships among initial features, hidden features, and labels.
To this end, we further introduce four additional Gram-type statistics, whose analytical expressions are given as follows:
\begin{equation}
\label{eq:app-Implementation-2}
    \widetilde{\mathbf{M}}_{l}^{t}
    =
    (\widetilde{\mathbf{F}}_{l}^{t})^{\top}
    \widetilde{\mathbf{Y}}^{t},
    \quad
    \widetilde{\mathbf{\Pi}}_{l}^{t}
    =
    (\widetilde{\mathbf{F}}_{l}^{t})^{\top}
    \widetilde{\mathbf{\Phi}}_{0}^{t},
    \quad
    \widetilde{\mathbf{\Psi}}_{l,i}^{t}
    =
    (\widetilde{\mathbf{F}}_{l}^{t})^{\top}
    \widetilde{\mathbf{F}}_{i}^{t},
    \quad
    \widetilde{\mathbf{\Gamma}}_{l}^{t}
    =
    (\widetilde{\mathbf{F}}_{l}^{t})^{\top}
    \widetilde{\mathbf{\Phi}}_{l}^{t}.
\end{equation}

Among these eight statistics, $\widetilde{\mathbf{G}}_{l}^{t}$, $\widetilde{\mathbf{H}}_{l}^{t}$, $\widetilde{\mathbf{N}}_{l}^{t}$, $\widetilde{\mathbf{M}}_{l}^{t}$, $\widetilde{\mathbf{\Pi}}_{l}^{t}$, and $\widetilde{\mathbf{\Psi}}_{l,i}^{t}$ are the six core Gram-type statistics maintained and recursively updated through our cross-recursive strategy, as illustrated in Fig.~\ref{fig:fig2}.
Specifically, depending on their relationships with the model weights, these statistics are updated through either phase-wise accumulation or layer-wise reconstruction.
Meanwhile, $\widetilde{\mathbf{\Gamma}}_{l}^{t}$ and $\widetilde{\bm{\Upsilon}}_{l}^{t}$ serve as two intermediate statistics derived from the maintained quantities.
By recursively updating these statistics and deriving the required intermediate quantities, the server can compute the analytic classifiers and transformation matrices in closed form, thereby continually constructing the multi-layer analytic network without revisiting historical data.

\subsection{Detailed Procedures on the Client Side}
\label{sec:implementation-worker}

Here, we further introduce the detailed computations performed on the client side.
Specifically, for an arbitrary residual branch $l\in[0,L-1]$ within service phase $t$, each client $k$ first performs feature extraction using its locally sensed data and constructs the corresponding feature representations, as shown in \eqref{eq:app-Implementation-3}.
Based on the extracted initial features $\mathbf{\Phi}_{0}^{t,k}$, the client then computes the corresponding hidden feature matrix $\mathbf{F}_{l}^{t,k}$ using \eqref{eq:app-Implementation-4}.
\begin{equation}
\label{eq:app-Implementation-3}
\mathbf{\Phi}_{0}^{t,k} = \sigma(\text{Backbone}(\mathbf{X}^{t,k}, \mathbf{\Theta}) \mathbf{A}), \quad
\mathbf{\Phi}_{l}^{t,k} = \mathbf{\Phi}_{l-1}^{t,k} + \mathbf{F}_{l-1}^{t,k}\mathbf{\Omega}_{l}^{t},
\ \forall l \in [1,L].
\end{equation}
\begin{equation}
\label{eq:app-Implementation-4}
\mathbf{F}_{l}^{t,k} = \sigma(\mathbf{\Phi}_{0}^{t,k} \mathbf{B}_l), \ \forall l \in [0,L-1].
\end{equation}

Upon obtaining the feature representations, each client constructs the local Gram-type statistics required for subsequent global model updates.
Instead of transmitting the original feature matrices or raw observations, clients compute inner-product statistics that capture the essential relationships among hidden features, initial features, and labels.
Specifically, each client $k$ constructs $\mathbf{N}_{l}^{t,k}$, $\mathbf{M}_{l}^{t,k}$, $\mathbf{\Pi}_{l}^{t,k}$, and $\{\mathbf{\Psi}_{l,i}^{t,k}\}_{i=0}^{l-1}$ as follows:
\begin{equation}
\label{eq:app-Implementation-5}
\mathbf{N}_{l}^{t,k} = (\mathbf{F}_{l}^{t,k})^{\top}\mathbf{F}_{l}^{t,k}, \quad \mathbf{M}_{l}^{t,k} = (\mathbf{F}_{l}^{t,k})^{\top}\mathbf{Y}^{t,k}, \quad \mathbf{\Pi}_{l}^{t,k} = (\mathbf{F}_{l}^{t,k})^{\top}\mathbf{\Phi}_{0}^{t,k}, \quad \forall l \in [0,L-1].
\end{equation}
\begin{equation}
\label{eq:app-Implementation-5-2}
\mathbf{\Psi}_{l,i}^{t,k} = (\mathbf{F}_{l}^{t,k})^{\top}\mathbf{F}_{i}^{t,k}, \quad \forall i \in [0,l),\ \forall l \in [1,L-1].
\end{equation}
Among these statistics, $\mathbf{N}_{l}^{t,k}$ captures the correlations within the hidden features, while $\mathbf{M}_{l}^{t,k}$ and $\mathbf{\Pi}_{l}^{t,k}$ describe their relationships with the local labels and initial features, respectively.
Furthermore, $\mathbf{\Psi}_{l,i}^{t,k}$ captures the correlations between hidden features from different layers, which are required for the subsequent layer-wise reconstruction.

Notably, the initial layer $l=0$ is a special case.
Since no preceding hidden features exist at this layer, the computation of $\mathbf{\Psi}_{l,i}^{t,k}$ is omitted.
Instead, to initialize the recursive construction of the global Gram-type statistics, each client additionally computes the following two local statistics based on the initial feature matrix:
\begin{equation}
\label{eq:app-Implementation-6}
\mathbf{H}_{0}^{t,k}
=
(\mathbf{\Phi}_{0}^{t,k})^{\top}\mathbf{Y}^{t,k},
\quad
\mathbf{G}_{0}^{t,k}
=
(\mathbf{\Phi}_{0}^{t,k})^{\top}\mathbf{\Phi}_{0}^{t,k}.
\end{equation}
These two statistics provide the initial-layer information required to update $\widetilde{\mathbf{H}}_{0}^{t}$ and $\widetilde{\mathbf{G}}_{0}^{t}$, which further serve as the starting point for the subsequent layer-wise reconstruction on the server side.

Finally, each client uploads its local Gram-type statistics to the server.
Specifically, for $l=0$, the uploaded statistics include $\mathbf{N}_{0}^{t,k}$, $\mathbf{M}_{0}^{t,k}$, $\mathbf{\Pi}_{0}^{t,k}$, $\mathbf{H}_{0}^{t,k}$, and $\mathbf{G}_{0}^{t,k}$.
For $l\in[1,L-1]$, the uploaded statistics include $\mathbf{N}_{l}^{t,k}$, $\mathbf{M}_{l}^{t,k}$, $\mathbf{\Pi}_{l}^{t,k}$, and $\{\mathbf{\Psi}_{l,i}^{t,k}\}_{i=0}^{l-1}$.
In this way, clients only need to contribute local statistics, while the server can reconstruct the required global information through the cross-recursive strategy, as detailed in the next section.

\subsection{Detailed Procedures on the Server Side}
\label{sec:implementation-platform}

Here, we further detail the procedures performed on the server side to implement the cross-recursive strategy.
After receiving the local Gram-type statistics uploaded by participating clients, the server recursively updates the global statistics through two complementary operations: phase-wise recursion update and layer-wise recursion update.
Based on these updated statistics, the server subsequently computes the global analytic classifiers and learnable transformation matrices in closed form, thereby continually constructing the multi-layer analytic network without accessing historical raw data.

Before performing the recursive updates, the server first initializes all six maintained global Gram-type statistics as zero matrices at phase $t=0$.
This initialization provides the starting point for incorporating successive client contributions throughout the service construction process:
\begin{equation}
\label{eq:app-Implementation-init}
\widetilde{\mathbf{G}}_{l}^{0} = \mathbf{0},
\ \
\widetilde{\mathbf{H}}_{l}^{0} = \mathbf{0},
\ \
\widetilde{\mathbf{N}}_{l}^{0} = \mathbf{0},
\ \
\widetilde{\mathbf{M}}_{l}^{0} = \mathbf{0},
\ \
\widetilde{\mathbf{\Pi}}_{l}^{0} = \mathbf{0},
\ \
\widetilde{\mathbf{\Psi}}_{l,i}^{0} = \mathbf{0},
\ \
\forall l\in[0,L].
\end{equation}

Then, the server proceeds to recursively update the global Gram-type statistics using the newly received client contributions.
As discussed in Section~\ref{Section-3.2}, statistics involving only the hidden features $\widetilde{\mathbf{F}}_{l}^{t}$, initial features $\widetilde{\mathbf{\Phi}}_{0}^{t}$, and labels $\widetilde{\mathbf{Y}}^{t}$ can be directly accumulated across service phases.
This is because their corresponding feature mappings remain independent of the updated model weights, allowing previously accumulated statistics to be reused without recomputing historical features.
Therefore, as highlighted by the \myblueA{\textbf{\textit{blue blocks}}} in Fig.~\ref{fig:fig2}, the statistics $\widetilde{\mathbf{H}}_{0}^{t}$, $\widetilde{\mathbf{G}}_{0}^{t}$, $\widetilde{\mathbf{N}}_{l}^{t}$, $\widetilde{\mathbf{M}}_{l}^{t}$, $\widetilde{\mathbf{\Pi}}_{l}^{t}$, and $\widetilde{\mathbf{\Psi}}_{l,i}^{t}$ can be recursively updated by adding the current clients' local statistics to the previously accumulated global statistics, as follows:
\begin{equation}
\label{eq:app-Implementation-7}
\begin{aligned}
\begin{cases}
\mathbf{\widetilde H}_{0}^{t}
= [\mathbf{\widetilde H}_{0}^{t-1}, \mathbf{0}_{d_{\Phi}\times C'_t}]
+ \sum\nolimits_{k \in S_t}\mathbf{H}_{0}^{t,k}; \\
\mathbf{\widetilde G}_{0}^{t}
= \mathbf{\widetilde G}_{0}^{t-1}
+ \sum\nolimits_{k \in S_t}\mathbf{G}_{0}^{t,k};\\
\mathbf{\widetilde N}_{l}^{t} =
\mathbf{\widetilde N}_{l}^{t-1} + \sum\nolimits_{k \in S_t}\mathbf{N}_{l}^{t,k}, \quad \forall l \in [0,L-1]; \\
\mathbf{\widetilde M}_{l}^{t} =
[\mathbf{\widetilde M}_{l}^{t-1}, \mathbf{0}_{d_\text F \times C'_t}] + \sum\nolimits_{k \in S_t}\mathbf{M}_{l}^{t,k}, \quad \forall l \in [0,L-1]; \\
\mathbf{\widetilde \Pi}_{l}^{t} =
\mathbf{\widetilde \Pi}_{l}^{t-1} + \sum\nolimits_{k \in S_t}\mathbf{\Pi}_{l}^{t,k}, \quad \forall l \in [0,L-1]; \\
\mathbf{\widetilde \Psi}_{l,i}^{t} =
\mathbf{\widetilde \Psi}_{l,i}^{t-1} + \sum\nolimits_{k \in S_t}\mathbf{\Psi}_{l,i}^{t,k}, \quad \forall i \in [0,l), \forall l \in [1,L-1];
\end{cases}
\end{aligned}
\end{equation}
Here, $C'_t = C_t - C_{t-1}$ denotes the number of newly introduced classes in phase $t$.
Accordingly, the zero matrices $\mathbf{0}_{d_{\Phi}\times C'_t}$ and $\mathbf{0}_{d_\text{F}\times C'_t}$ are concatenated to the previously accumulated statistics $\widetilde{\mathbf{H}}_{0}^{t-1}$ and $\widetilde{\mathbf{M}}_{l}^{t-1}$, respectively, to align their dimensions with the expanded class space.
This operation preserves the accumulated information of existing classes while accommodating new service demands.

But the above phase-wise accumulation cannot be directly applied to statistics involving the learned deep features $\widetilde{\mathbf{\Phi}}_{l}^{t}$ with $l\ge1$.
As explained in Section~\ref{Section-3.2}, updating the weights of preceding layers also changes the intermediate representations of historical samples, making their previously accumulated statistics inconsistent with the updated model.
To address this issue, the server reconstructs the corresponding global statistics using the feature relationships between adjacent layers, thereby incorporating the updated model weights without revisiting historical data.
Specifically, as highlighted by the {\color{myred}\textbf{\textit{red blocks}}} in Fig.~\ref{fig:fig2}, $\widetilde{\mathbf{H}}_{l}^{t}$ and $\widetilde{\mathbf{G}}_{l}^{t}$ with $l\ge1$ are recursively updated in a layer-wise manner as follows:
\begin{equation}
\label{eq:app-Implementation-8}
\begin{cases}
\mathbf{\widetilde H}_{l}^{t}
= \mathbf{\widetilde H}_{l-1}^{t}
+ (\mathbf{\Omega}_{l}^{t})^{\top}\mathbf{\widetilde M}_{l-1}^{t},
\ \forall l \in [1,L]; \\
\mathbf{\widetilde G}_{l}^{t}
= \mathbf{\widetilde G}_{l-1}^{t}
+ (\mathbf{\widetilde \Gamma}_{l-1}^{t})^{\top}\mathbf{\Omega}_{l}^{t}
+ (\mathbf{\Omega}_{l}^{t})^{\top}\mathbf{\widetilde \Gamma}_{l-1}^{t} 
+ (\mathbf{\Omega}_{l}^{t})^{\top}\mathbf{\widetilde N}_{l-1}^{t}\mathbf{\Omega}_{l}^{t},
\ \forall l \in [1,L];
\end{cases}
\end{equation}

Notably, the layer-wise reconstruction of $\widetilde{\mathbf{G}}_{l}^{t}$ additionally requires the intermediate Gram-type statistic $\widetilde{\mathbf{\Gamma}}_{l}^{t}$, defined as $(\widetilde{\mathbf{F}}_{l}^{t})^{\top}\widetilde{\mathbf{\Phi}}_{l}^{t}$.
Since the server cannot directly access the global feature matrices of historical samples, this statistic must also be recovered from the available information.
Specifically, by expanding the learned features through their residual connections, the server can derive $\widetilde{\mathbf{\Gamma}}_{l}^{t}$ from the recursively accumulated statistics $\widetilde{\mathbf{\Pi}}_{l}^{t}$ and $\{\widetilde{\mathbf{\Psi}}_{l,i}^{t}\}_{i=0}^{l-1}$, together with the current transformation matrices, as follows:
\begin{equation}
\label{eq:app-Implementation-9}
\mathbf{\widetilde{\Gamma}}_{0}^{t} 
    = (\mathbf{\widetilde F}_{0}^{t} )^{\top}\mathbf{\widetilde\Phi}_{0}^{t}   
    = \mathbf{\widetilde \Pi}_{0}^{t}, \quad
    \mathbf{\widetilde{\Gamma}}_{l}^{t}  =\mathbf{\widetilde \Pi}_{l}^{t} + \sum\nolimits_{i=0}^{l-1} \mathbf{\widetilde \Psi}_{l,i}^{t}\mathbf{\Omega}_{i+1}^{t}, \ \forall l \in [1,L-1].
\end{equation}

Through the preceding phase-wise and layer-wise recursions, the server obtains the global statistics required for the subsequent closed-form computations.
As illustrated by the {\color{green!50!black}\textbf{\textit{green blocks}}} in Fig.~\ref{fig:fig2}, these statistics are then used to derive the analytic classifiers and learnable transformation matrices, as detailed below.

Specifically, the server first computes the analytic classifier $\mathbf{W}_{l}^{t}$ for the current network layer.
As discussed in Section~\ref{Section-3.2}, its closed-form solution depends only on the global Gram-type statistics $\widetilde{\mathbf{G}}_{l}^{t}$ and $\widetilde{\mathbf{H}}_{l}^{t}$, which have already been obtained through the preceding recursive updates.
Therefore, the server can directly compute the analytic classifier without accessing the original features or labels, as follows:
\begin{equation}
\label{eq:app-Implementation-10}
    \mathbf{W}_{l}^{t}
    = ( \widetilde{\mathbf{G}}_{l}^{t} +
    \lambda\mathbf{I} )^{-1}
    \widetilde{\mathbf{H}}_{l}^{t}.
\end{equation}

For $l<L$, after obtaining the analytic classifier $\mathbf{W}_{l}^{t}$, the server further derives the intermediate Gram-type statistic $\widetilde{\bm{\Upsilon}}_{l}^{t}$, which captures the correlations between the hidden features and the current prediction residuals.
Although its definition involves the residual matrix $\widetilde{\mathbf{R}}_{l}^{t}$, this statistic can be directly recovered from the previously obtained $\widetilde{\mathbf{M}}_{l}^{t}$, $\widetilde{\mathbf{\Gamma}}_{l}^{t}$, and $\mathbf{W}_{l}^{t}$.
Specifically, we can obtain:
\begin{equation}
\label{eq:app-Implementation-11}
    \widetilde{\bm{\Upsilon}}_{l}^{t}
    = (\widetilde{\mathbf{F}}_{l}^{t})^{\top}
    \widetilde{\mathbf{R}}_{l}^{t}
    = \widetilde{\mathbf{M}}_{l}^{t}
    - \widetilde{\mathbf{\Gamma}}_{l}^{t}
    \mathbf{W}_{l}^{t}.
\end{equation}

In this way, the server obtains the required residual-related information without explicitly reconstructing the historical residual matrices.
Together with the updated global statistic $\widetilde{\mathbf{N}}_{l}^{t}$ and the analytic classifier $\mathbf{W}_{l}^{t}$, the server can subsequently determine the optimal learnable transformation matrix $\mathbf{\Omega}_{l+1}^{t}$ through closed-form computations.
Specifically, based on the analytical solution derived in Section~\ref{sec:recalm-learning}, we can compute $\mathbf{\Omega}_{l+1}^{t}$ as follows:
\begin{equation}
\label{eq:app-Implementation-12}
    \mathbf{\Omega}_{l+1}^{t}
    = \mathbf{V}_{l}^{t}
    \left[
    \left(
    \mathbf{V}_{l}^{t\top}
    \widetilde{\bm{\Upsilon}}_{l}^{t}
    \mathbf{W}_{l}^{t\top}
    \mathbf{U}_{l}^{t}
    \right)
    \oslash
    \mathbf{T}_{l}^{t}
    \right]
    \mathbf{U}_{l}^{t\top},
    \quad
    \mathbf{T}_{l}^{t}
    = \gamma\mathbf{1}
    + \operatorname{diag}(\mathbf{\Lambda}_{l,\mathrm{H}}^{t})
    \otimes
    \operatorname{diag}(\mathbf{\Lambda}_{l,\mathrm{W}}^{t}).
\end{equation}

Here, $\widetilde{\mathbf{N}}_{l}^{t}
= \mathbf{V}_{l}^{t}
\mathbf{\Lambda}_{l,\mathrm{H}}^{t}
\mathbf{V}_{l}^{t\top}$
and
$\mathbf{W}_{l}^{t}\mathbf{W}_{l}^{t\top}
= \mathbf{U}_{l}^{t}
\mathbf{\Lambda}_{l,\mathrm{W}}^{t}
\mathbf{U}_{l}^{t\top}$
represent the corresponding spectral decompositions.
The resulting transformation matrix $\mathbf{\Omega}_{l+1}^{t}$ enables the residual feature correction required for constructing the subsequent network layer.

Finally, for $l<L$, the server transmits the computed $\mathbf{\Omega}_{l+1}^{t}$ to the participating clients, allowing them to construct the next-layer features according to \eqref{eq:app-Implementation-3}.
The same procedure is then recursively executed across network layers and successive service phases, enabling ReCALM to continually update its multi-layer analytic network while incorporating all previously accumulated client contributions.
After completing the final service phase $T$, the server obtains the global parameters $\{\mathbf{\Omega}_{l}^{T}\}_{l=1}^{L}$ and $\{\mathbf{W}_{l}^{T}\}_{l=0}^{L}$.

For completeness, the training and inference procedures of our ReCALM are summarized in the Appendix.
In addition, we provide a detailed theoretical validation of the correctness of our distributed implementation, which is also presented in the Appendix.

\subsection{Flexibility for Dynamic Client Participation and Asynchronous Updates}
\label{sec:worker-participation}

In the preceding sections, we have introduced the implementation of ReCALM under a synchronous setting, where clients participating in each service phase first compute and upload their local Gram-type statistics, and the server then aggregates these contributions to update the global service model.
However, this synchronous procedure is mainly adopted for clarity of presentation rather than a strict requirement of ReCALM.
In practical crowdsourced sensing scenarios, clients may dynamically join or leave the system, temporarily disconnect, or contribute newly sensed data at different times.
Therefore, it is desirable for the server to continually update its service model without waiting for all clients to participate simultaneously.

\clearpage

In fact, ReCALM can naturally accommodate such dynamic participation and asynchronous updates through its cross-recursive strategy.
The key insight is that the server updates its model using accumulated Gram-type statistics, rather than relying on a fixed set of clients or their participation order.
More importantly, as established by the heterogeneity invariance property in Theorem~1, different aggregation schedules yield the same final analytic solution when the same set of client contributions is incorporated.
This enables ReCALM to flexibly construct its multi-layer analytic network under both synchronous and asynchronous settings.
We further elaborate on this flexibility from the following three aspects.

\begin{itemize}
    \item \textbf{Dynamic Client Participation.}
    In practical crowdsourced sensing systems, clients may participate dynamically due to differences in availability, sensing opportunities, computational capabilities, and communication conditions.
    Some clients may join the system for the first time, while others may return after previous participation to contribute newly collected observations.
    Moreover, certain clients may temporarily disconnect or fail to upload their local statistics within the current service phase.
    Fortunately, ReCALM does not require clients to retain historical data or participate continuously.
    Instead, each newly received contribution can be uniformly treated as a \emph{virtual client}, regardless of whether it originates from a newly recruited or returning client.
    If certain clients fail to contribute on time, the server can simply proceed with the statistics received from other participating clients.
    When delayed or returning clients subsequently provide additional data, their contributions can be incorporated through the same recursive update procedure.
    {\textit{Consequently, the absence or delayed participation of individual clients does not prevent the server from continually updating its service model.}}

    \item \textbf{Compatibility with Asynchronous Aggregation.}
    Beyond dynamic client participation, ReCALM can support asynchronous aggregation, where client contributions arrive at different times and may not follow the original service-phase boundaries.
    For example, the server may have incorporated contributions associated with a newly introduced service demand, while some clients from earlier phases are still uploading their local statistics.
    Nevertheless, since the cross-recursive strategy updates the model through accumulated Gram-type statistics, the server does not need to strictly synchronize all client contributions before proceeding.
    {\textit{In practice, this can be implemented in two ways.}}
    First, the server can buffer the received local Gram-type statistics and aggregate them once a sufficient number of client contributions have arrived.
    Second, the server can treat each asynchronously received contribution as an individual service phase and directly incorporate its statistics into the maintained global Gram-type statistics.
    Both approaches employ the same cross-recursive update procedure, allowing the server to accommodate different contribution schedules without accessing historical raw data.

    \item \textbf{Unified Implementation and Invariant Results.}
    Whether clients participate synchronously or asynchronously, the core server-side computation remains unchanged.
    Specifically, the server incorporates newly received local Gram-type statistics into the maintained global statistics and reconstructs the multi-layer analytic network through closed-form solutions.
    The essential difference between synchronous and asynchronous settings lies in how client contributions are grouped into service phases and when the corresponding updates are performed.
    As established by the heterogeneity invariance of ReCALM in Theorem~1, different partitions of the same complete dataset yield identical centralized analytic solutions under fixed feature mappings and consistent class indices.
    Therefore, provided that each contribution is incorporated exactly once, {\textit{different synchronous or asynchronous aggregation schedules yield the same final ReCALM model for the same client contributions, apart from numerical precision.}}
    Notably, if a client's contribution is permanently unavailable, the server can continue updating its model using the received data, while centralized equivalence holds for these accumulated contributions.
    This flexibility is valuable for crowdsourced continual service construction, where client participation and communication conditions may change throughout the service lifetime.
\end{itemize}

\section{Experiments}
\label{sec:experiments}

In this section, we compare ReCALM with representative methods across three established datasets.
We then analyze its deep representation learning capability, computational efficiency, and sensitivity to parameter choices and architectural components.
To clarify the objectives of these experiments, we address the following four Research Questions (RQs):

\begin{itemize}
    \item[\textbf{RQ1.}] How does ReCALM's accuracy compare with existing methods across client and service-phase partitions?
    \item[\textbf{RQ2.}] How do residual analytic layers improve fitting and feature separability as network depth increases?
    \item[\textbf{RQ3.}] What accuracy--efficiency trade-offs does ReCALM provide across network depths?
    \item[\textbf{RQ4.}] How do hyperparameter choices and individual architectural components affect ReCALM's accuracy?
\end{itemize}

\subsection{Experimental Setup}
\label{sec:Experimental Setup}

Here, we briefly introduce the experimental setup for systematically evaluating ReCALM in crowdsourced continual service construction, including the datasets, competing baselines, implementation configurations, and evaluation metrics.
Further details on the experiments are provided in the Appendix.

\textbf{Datasets.}
We conduct experiments on CIFAR-100~\cite{dataset_1}, Tiny-ImageNet~\cite{dataset_2}, and ImageNet-R~\cite{dataset_ImageNet-R}.
To simulate crowdsourced continual service construction with evolving service demands and distributed client contributions, we partition each dataset across service phases and clients.
Following AFCL~\cite{AFCL}, we adopt the Si-Blurry setting~\cite{Si-Blurry} to partition each dataset into $T=5$ or $T=10$ service phases, mixing previously observed and new classes in the incremental data stream~\cite{GACL}.
We use a disjoint-class proportion of $r_\text{D}=50\%$ and a blurry-data redistribution proportion of $r_\text{B}=10\%$.
Within each phase, the data are distributed among five clients using a Dirichlet distribution~\cite{Dirichlet}, simulating heterogeneous sensing contributions across clients.
Its concentration parameter $\alpha$ controls differences in clients' class distributions: we use $\alpha\in\{0.1,0.2\}$ for CIFAR-100 and Tiny-ImageNet, and $\alpha\in\{0.5,1.0\}$ for ImageNet-R.
All methods use identical partitions within each setting.

\textbf{Baselines.}
We compare ReCALM against eleven baselines from three groups.
The FL group includes FedAvg~\cite{fedavg}, FedProx~\cite{FedProx}, and FedDyn~\cite{Feddyn}.
The FL+CL group combines continual-learning techniques with federated training: FedIcaRL~\cite{iCaRL}, FedLwF~\cite{LwF}, and FedEwc~\cite{Ewc}.
The FCL group includes FedMGP~\cite{FCL-2}, LGA~\cite{LGA}, FedCBC~\cite{FCL-3}, TagFCL~\cite{tagfcl}, and AFCL~\cite{AFCL}.
To ensure fair comparisons under distributed and incremental data settings, we follow the experimental protocol of AFCL~\cite{AFCL}, which aligns with crowdsourced continual service construction.
Notably, under matched feature mappings and regularization, AFCL~\cite{AFCL} and CALM~\cite{CALM} share the same single-layer linear analytic classifier.
Therefore, we regard them as equivalent baselines and use AFCL for comparison in this federated setting.
Following AFL~\cite{AFL}, and CALM~\cite{CALM}, we adopt ResNet-18~\cite{ResNet} as the unified backbone for all methods.
Additionally, we employ two stronger ViT-B/16 backbones pre-trained on ImageNet-1K and ImageNet-21K, respectively, to investigate ReCALM's performance with stronger feature extractors.

\textbf{Metrics.}
To evaluate recognition performance, we use the average accuracy of the final service model on the complete test set covering all recognition classes as our primary metric.
We report phase-wise accuracy on the cumulative test set to examine continual service construction, together with computational costs to evaluate efficiency.
Furthermore, to assess the effectiveness of learned deep representations, we consider training accuracy and three complementary feature-separability metrics, namely compactness-separation ratio, inverse fisher score, and discriminative measure, whose detailed definitions are provided in the Appendix.

\textbf{Implementations.}
We adopt GELU as the default activation function, projection dimensions $d_\Phi=d_\text{F}=1024$, and regularization parameters $\lambda=1$ and $\gamma=0.01$.
We selected these default settings based on preliminary experiments using only 10\% of the training data, without systematic hyperparameter tuning.
Although our sensitivity analysis reveals that better-performing configurations exist, ReCALM's robustness and invariance reduce the need for extensive hyperparameter tuning.
This further highlights the practical advantages of ReCALM.
\clearpage

\subsection{RQ1: Performance Comparison}
\label{sec:exp:rq1}

\begin{table*}[t]
\centering
\caption{
Performance comparison of average accuracy among our ReCALM and the baselines.
The parameters $T$ and $\alpha$ specify the number of service phases and the distribution of classes across clients, respectively.
The {\color{myblue}\textbf{bold}} and {\color{my2}\underline{underlined}} results indicate the best and second-best performance, respectively.
} 
\renewcommand{\arraystretch}{1.15}
\resizebox{1.0\textwidth}{!}{
\begin{NiceTabular}{@{}cc|cccc|cccc|cccc@{}}
\toprule[1.2pt]
\multicolumn{2}{c}{\multirow{4}{*}{{Method}}} &
  \multicolumn{4}{c}{CIFAR-100} &
  \multicolumn{4}{c}{Tiny-ImageNet} &
  \multicolumn{4}{c}{ImageNet-R} \\ \cmidrule[0.4pt](l){3-14} 
  \multicolumn{2}{c}{} &
  \multicolumn{2}{c|}{$T=5$} &
  \multicolumn{2}{c}{$T=10$} &
  \multicolumn{2}{c|}{$T=5$} &
  \multicolumn{2}{c}{$T=10$} &
  \multicolumn{2}{c|}{$T=5$} &
  \multicolumn{2}{c}{$T=10$} \\ 
  \cmidrule[0.4pt](l){3-14} 
\multicolumn{2}{c}{} &
  $\alpha=0.1$ &
  \multicolumn{1}{c|}{$\alpha=0.2$} &
  $\alpha=0.1$ &
  \multicolumn{1}{c}{$\alpha=0.2$} &
  $\alpha=0.1$ &
  \multicolumn{1}{c|}{$\alpha=0.2$} &
  $\alpha=0.1$ &
  $\alpha=0.2$ &
  $\alpha=0.5$ &
  \multicolumn{1}{c|}{$\alpha=1.0$} &
  $\alpha=0.5$ &
  $\alpha=1.0$  \\ \midrule[0.4pt]
\multirow{3}{*}{FL}  & \multicolumn{1}{|c}{FedAvg} & 4.96\% & \multicolumn{1}{c|}{\; 6.35\%} & \; 4.15\% & \multicolumn{1}{c|}{ \; 4.98\%} & \; 1.85\% & \multicolumn{1}{c|}{ \; 2.24\%} & \; 1.32\% & \; 1.47\% & \; 1.41\% & \multicolumn{1}{c|}{ \; 1.73\%} & \; 1.47\% & \; 1.65\%  \\
 & \multicolumn{1}{|c}{FedProx}  & 10.01\% & \multicolumn{1}{c|}{12.56\%} & \; 7.92\% & 10.45\% & \; 6.73\% & \multicolumn{1}{c|}{ \; 7.85\%} & \; 4.83\% & \; 6.27\% & \; 9.79\% & \multicolumn{1}{c|}{12.46\%} & \; 6.58\% & \; 8.65\%  \\
& \multicolumn{1}{|c}{FedDyn}  & 11.50\% & \multicolumn{1}{c|}{13.64\%} & \; 9.01\% & 10.72\% & \; 6.65\% & \multicolumn{1}{c|}{14.66\%} & \; 5.37\% & 11.49\% & 13.41\% & \multicolumn{1}{c|}{14.81\%} & 11.95\% & 13.48\%  \\
\midrule[0.4pt]
{\multirow{3}{*}{FL+CL}} & \multicolumn{1}{|c}{FedIcaRL} &  4.05\% & \multicolumn{1}{c|}{ \; 5.52\%} & \; 3.64\% & \multicolumn{1}{c|}{ \; 3.93\%} & \; 2.05\% & \multicolumn{1}{c|}{ \; 2.41\%} & \; 1.84\% & \; 1.86\% & \; 2.03\% & \multicolumn{1}{c|}{ \; 2.22\%} & \; 1.57\% & \; 1.78\%  \\
 & \multicolumn{1}{|c}{FedLwF} & 8.37\% & \multicolumn{1}{c|}{11.11\%} & \; 4.18\% & \multicolumn{1}{c|}{ \; 5.63\%} & \; 3.13\% & \multicolumn{1}{c|}{ \; 3.62\%} & \; 2.70\% & \; 2.21\% & \; 3.13\% & \multicolumn{1}{c|}{ \; 3.19\%} & \; 2.17\% & \; 2.48\%  \\
 & \multicolumn{1}{|c}{FedEwc}  &  12.64\% & \multicolumn{1}{c|}{14.28\%} & \; 7.92\% & \multicolumn{1}{c|}{11.48\%} & \; 3.34\% & \multicolumn{1}{c|}{ \; 4.56\%} &  \; 2.63\% & \; 3.39\% & \; 4.27\% & \multicolumn{1}{c|}{ \; 4.71\%} & \; 3.21\% & \; 3.30\%  \\
\midrule[0.4pt]
\multirow{5}{*}{FCL} & \multicolumn{1}{|c}{FedMGP} &  23.87\% & \multicolumn{1}{c|}{26.75\%} & 15.21\% & \multicolumn{1}{c|}{16.17\%} & 12.53\% & \multicolumn{1}{c|}{14.82\%} & \; 9.60\% & 10.10\% & 18.23\% & \multicolumn{1}{c|}{18.69\%} & \; 8.75\% & \; 9.33\%  \\
&\multicolumn{1}{|c}{LGA}  & 39.09\% & \multicolumn{1}{c|}{40.21\%} & 31.78\% & {38.56\%} & 36.70\% & \multicolumn{1}{c|}{38.47\%} & 27.36\% & 29.33\% & 23.19\% & \multicolumn{1}{c|}{29.75\%} & 19.18\% & 21.59\%  \\
&\multicolumn{1}{|c}{FedCBC}  & 43.34\% & \multicolumn{1}{c|}{44.93\%} & 27.32\% & \multicolumn{1}{c|}{29.02\%} & 31.13\% & \multicolumn{1}{c|}{33.87\%} & 22.48\% & 24.62\% & 23.51\% & \multicolumn{1}{c|}{23.79\%} & 15.72\% & 14.14\%  \\
& \multicolumn{1}{|c}{TagFCL}  & {45.10\%} & \multicolumn{1}{c|}{46.64\%} & 36.66\% & 36.13\% & 40.94\% & \multicolumn{1}{c|}{44.98\%} & 29.60\% & 32.56\% & 34.44\% & \multicolumn{1}{c|}{38.01\%} & 24.76\% & 28.37\% \\
& \multicolumn{1}{|c}{AFCL}   & \color{my2}\underline{58.56\%} & \multicolumn{1}{c|}{\color{my2}\underline{{58.56\%}}} & \color{my2}\underline{58.56\%} & \multicolumn{1}{c|}{\color{my2}\underline{{58.56\%}}} & \color{my2}\underline{54.67\%} & \multicolumn{1}{c|}{\color{my2}\underline{{54.67\%}}} & \color{my2}\underline{54.67\%} & \color{my2}\underline{54.67\%} & \color{my2}\underline{38.22\%} & \multicolumn{1}{c|}{\color{my2}\underline{{38.22\%}}} & \color{my2}\underline{38.22\%} & \color{my2}\underline{38.22\%}  \\
\midrule[0.4pt]
\multicolumn{2}{c}{ReCALM $(L=5)$} & 65.02\% & \multicolumn{1}{c|}{65.02\%} & 65.02\% & \multicolumn{1}{c|}{65.02\%} & 61.05\% & \multicolumn{1}{c|}{61.05\%} & 61.05\% & 61.05\% & 46.37\% & \multicolumn{1}{c|}{46.37\%} & 46.37\% & 46.37\%  \\
\multicolumn{2}{c}{ReCALM $(L=10)$} & 66.75\% & \multicolumn{1}{c|}{66.75\%} & 66.75\% & \multicolumn{1}{c|}{66.75\%} & 62.23\% & \multicolumn{1}{c|}{62.23\%} & 62.23\% & 62.23\% & 47.15\% & \multicolumn{1}{c|}{47.15\%} & 47.15\% & 47.15\%  \\
\multicolumn{2}{c}{ReCALM $(L=20)$} & \color{myblue}\textbf{67.87}\% & \multicolumn{1}{c|}{\color{myblue}\textbf{{67.87\%}}} & \color{myblue}\textbf{67.87\%} & \multicolumn{1}{c|}{\color{myblue}\textbf{{67.87\%}}} & \color{myblue}\textbf{63.52\%} & \multicolumn{1}{c|}{\color{myblue}\textbf{{63.52\%}}} & \color{myblue}\textbf{63.52\%} & \color{myblue}\textbf{63.52\%} & \color{myblue}\textbf{48.05\%} & \multicolumn{1}{c|}{\color{myblue}\textbf{{48.05\%}}} & \color{myblue}\textbf{48.05\%} & \color{myblue}\textbf{48.05\%}  \\
\midrule[0.4pt]
\multicolumn{2}{c}{\textbf{Improvement }  $\bm{\uparrow}$} & \textbf{9.31\%} & \multicolumn{1}{c|}{\textbf{9.31\%}} & \textbf{9.31\%} & \textbf{9.31\%} & \textbf{8.85\%} & \multicolumn{1}{c|}{\textbf{8.85\%}} & \textbf{8.85\%} & \textbf{8.85\%} & \textbf{9.83\%} & \multicolumn{1}{c|}{\textbf{\; 9.83\%}} & \textbf{9.83\%} & \textbf{9.83\%}  \\
\bottomrule[1.2pt]
\end{NiceTabular}
}
\vspace{-0.4cm}
\label{table:performance}
\end{table*}

Here, we first conduct an overall performance comparison to evaluate the effectiveness of ReCALM in crowdsourced continual service construction.
Specifically, we compare our ReCALM with representative baselines under different client data distributions and service-phase partitions.
Table~\ref{table:performance} reports the final average accuracy across twelve dataset--partition settings.
With $L=20$, our ReCALM achieves 67.87\% on CIFAR-100, 63.52\% on Tiny-ImageNet, and 48.05\% on ImageNet-R, outperforming AFCL by 9.31\%, 8.85\%, and 9.83\%, respectively.
Even with only five residual layers, ReCALM achieves 65.02\%, 61.05\%, and 46.37\% on these datasets, surpassing all evaluated baselines.
Moreover, as further demonstrated in subsequent RQ2, increasing the depth beyond twenty layers can yield additional accuracy gains, indicating that this configuration does not fully exploit ReCALM's potential.
These results demonstrate the effectiveness of extending single-layer analytic network to a multi-layer analytic network.

Moreover, we can observe that our ReCALM achieves identical final accuracy across all evaluated client distributions and service-phase partitions for each dataset.
In contrast, the gradient-based baselines exhibit considerable performance degradation as client data become more heterogeneous or the number of service phases increases.
For example, on Tiny-ImageNet, TagFCL's accuracy decreases from 44.98\% to 40.94\% as $\alpha$ decreases from 0.2 to 0.1 at $T=5$, and further drops from 44.98\% to 32.56\% when $T$ increases from 5 to 10 at $\alpha=0.2$.
This is because these gradient-based methods suffer from conflicting updates across heterogeneous clients and catastrophic forgetting as new service demands arrive.
By contrast, our ReCALM remains unaffected by how the same data are partitioned across clients and service phases.
Although AFCL also exhibits this invariance, ReCALM achieves substantially higher recognition accuracy.
These experimental results further corroborate the heterogeneity-invariance property theoretically established in Theorem~1.

\begin{figure*}[t]
    \centering
    \includegraphics[width=\linewidth]{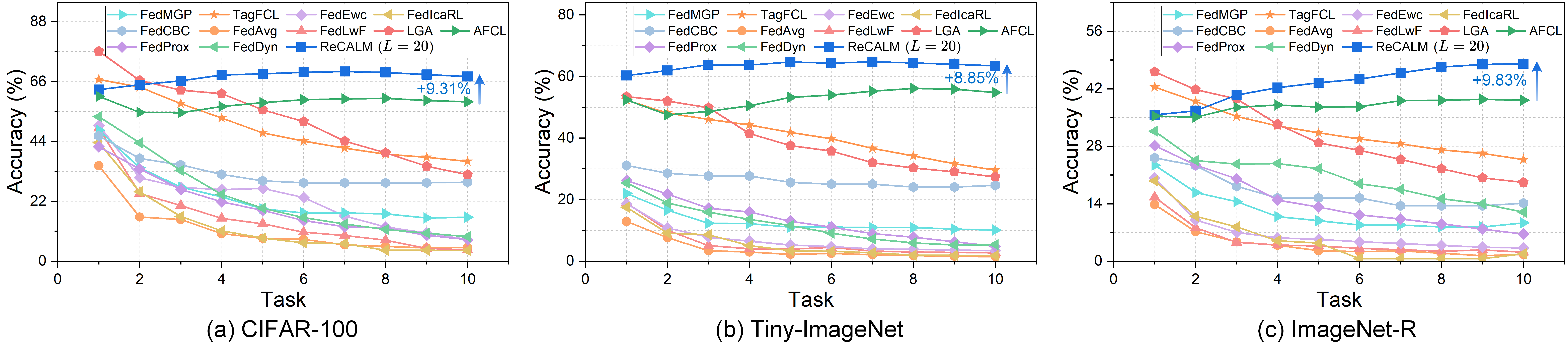}
    \vspace{-0.6cm}
    \caption{Average accuracy on the cumulative test set as service phases progress. }
    \label{fig:continual-accuracy}
    \vspace{-0.4cm}
\end{figure*}

Next, to further investigate the performance of different methods throughout continual service construction, we compare our ReCALM with the baselines across successive service phases.
Fig.~\ref{fig:continual-accuracy} presents their average accuracy on the cumulative test set as new service demands are progressively introduced.
As the number of service phases increases, the gradient-based methods generally exhibit considerable declines in accuracy, suggesting that they struggle to retain previously acquired knowledge during continual updates.
In contrast, our ReCALM maintains stable performance and even achieves further improvements as new data accumulate.
Moreover, ReCALM consistently outperforms AFCL throughout the entire service construction process, demonstrating the benefits of its learned deep representations.
\textbf{\textit{These results demonstrate that ReCALM achieves superior performance while maintaining stability throughout continual service construction.}}

\begin{figure*}[t]
\begin{minipage}[h]{0.329\linewidth}
		\centering
		\subfloat[CIFAR-100]{%
              \centering  
               \label{fig:Temporal-Invariance-CIFAR-100-migration-7}%
               \includegraphics[width=1.0\textwidth]{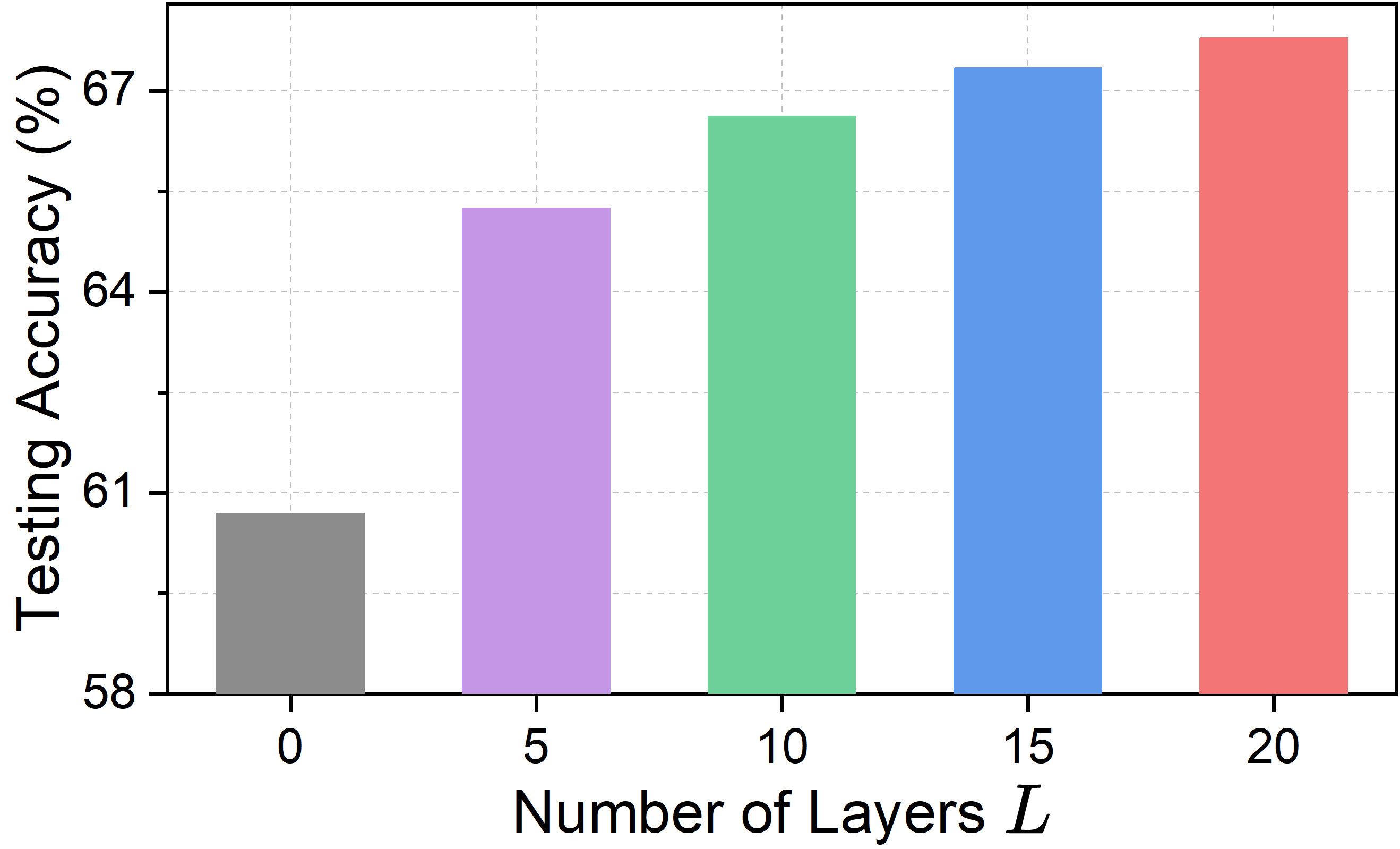}%
             } 
	\end{minipage}
	\hfill 
	\begin{minipage}[h]{0.329\linewidth}
		\centering
		\subfloat[Tiny-ImageNet]{%
              \centering  
               \label{fig:Temporal-Invariance-Tiny-ImageNet-free-migration-8}%
               \includegraphics[width=1.0\textwidth]{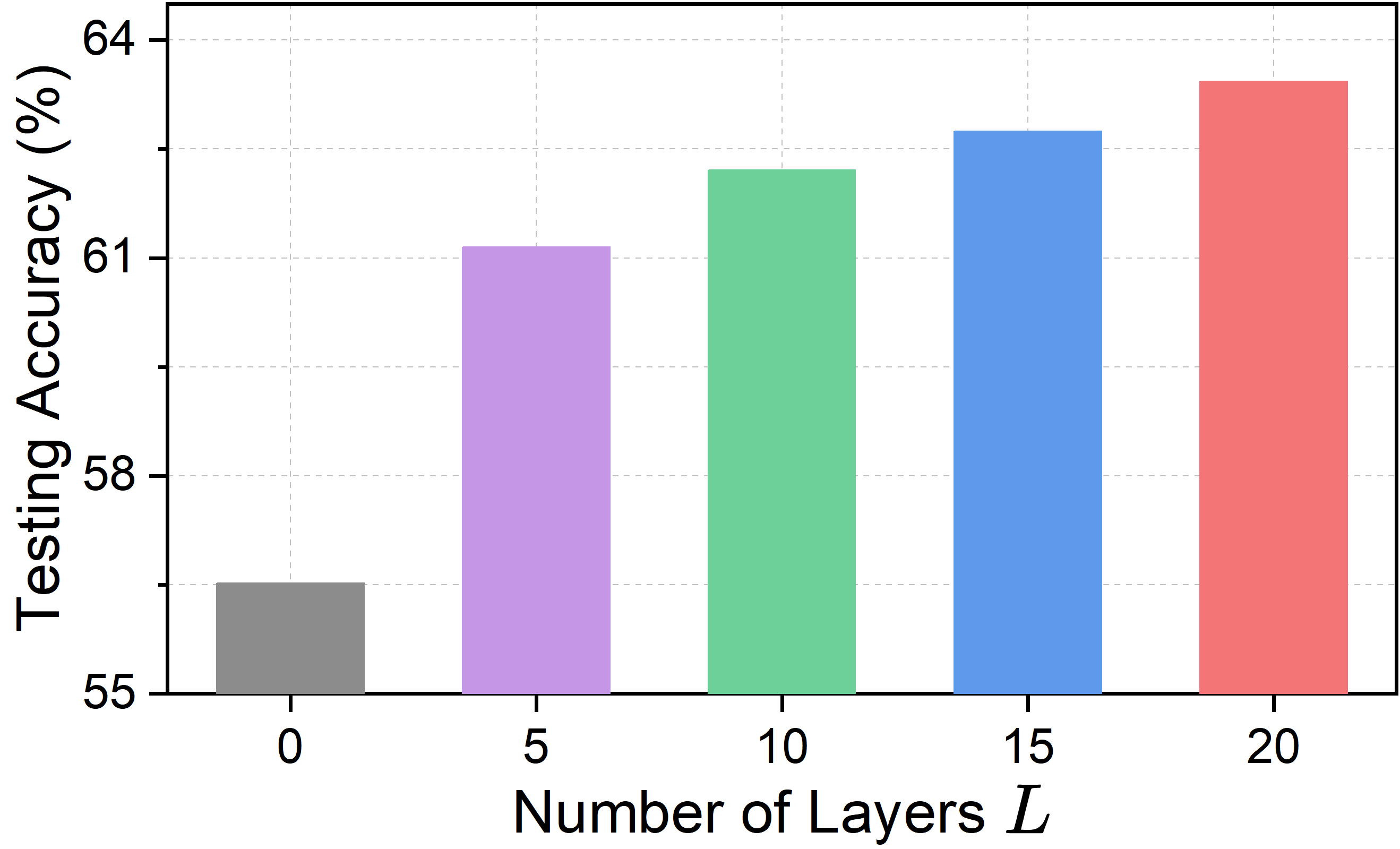}%
             } 
	\end{minipage}
    \hfill 
	\begin{minipage}[h]{0.329\linewidth}
		\centering
		\subfloat[ImageNet-R]{%
              \centering  
               \label{fig:Temporal-Invariance-Tiny-ImageNet-free-migration-9}%
               \includegraphics[width=1.0\textwidth]{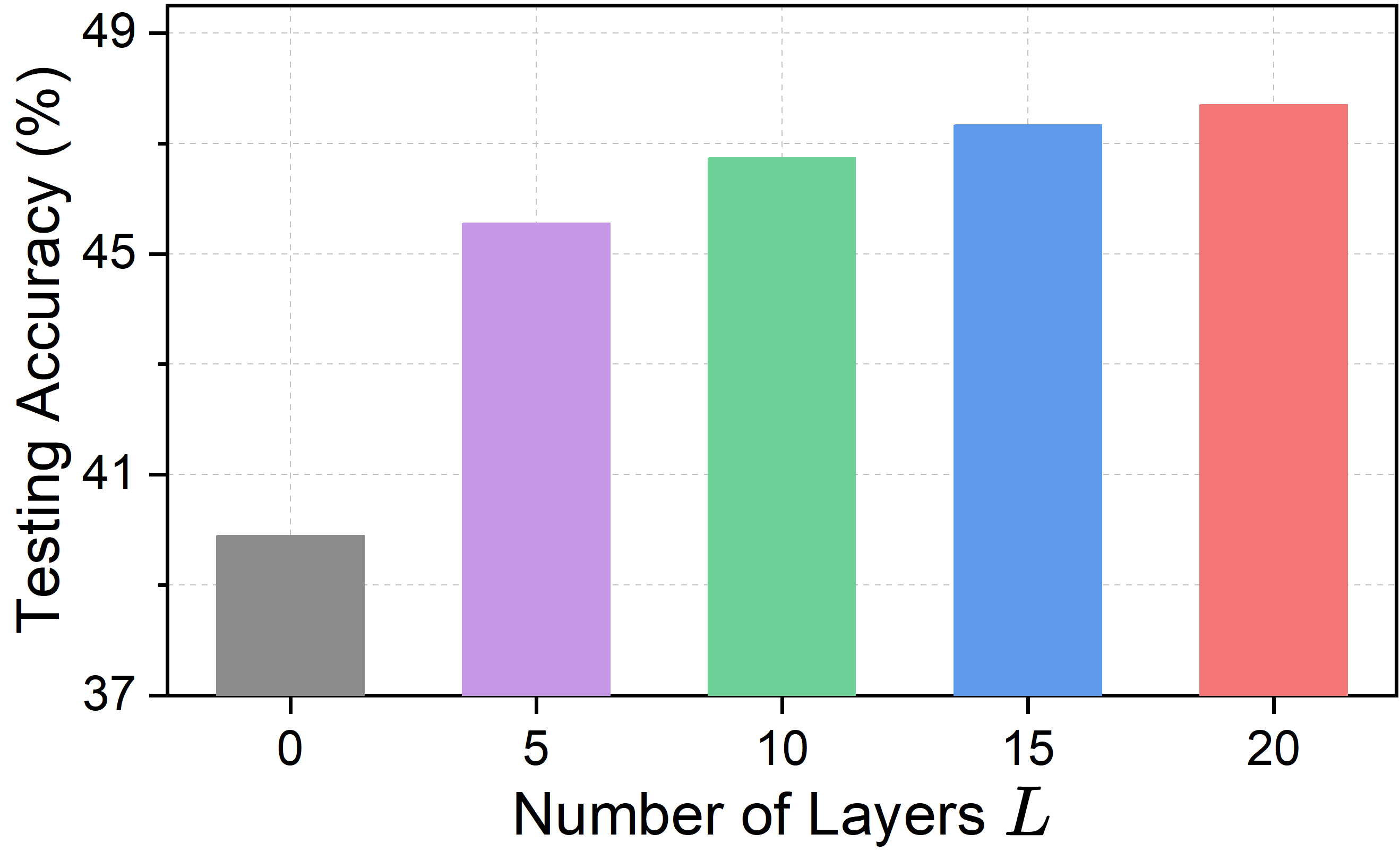}%
             } 
	\end{minipage}
 \vspace{-0.4cm}
 \caption{Testing accuracy of our ReCALM with different numbers of layers.}  %
 \vspace{-0.4cm}
 \label{app:fig:balance-2}
\end{figure*}

\subsection{RQ2: Deep Representation Learning}
\label{sec:exp:rq2}

Here, we further investigate the deep representation learning capability of our ReCALM, focusing on how its fitting ability and performance improve with increasing network depth.
Specifically, Fig.~\ref{app:fig:balance-2} presents the testing accuracy at different network depths across three datasets, while Table~\ref{app:tab:ViT} further reports the performance under different network depths and backbones on CIFAR-100.
More detailed training and testing results from $L=0$ to $L=50$ are provided in the Appendix.

As shown in Fig.~\ref{app:fig:balance-2}, the testing accuracy of our ReCALM progressively improves as network depth increases from $L=0$ to $L=20$ across all three datasets.
Specifically, the accuracy increases from 60.83\% to 67.88\% on CIFAR-100, from 56.76\% to 63.39\% on Tiny-ImageNet, and from 39.90\% to 47.70\% on ImageNet-R, corresponding to improvements of 7.05\%, 6.63\%, and 7.80\%, respectively.
Furthermore, the detailed results in the Appendix show that increasing the network depth to $L=50$ brings further gains on CIFAR-100 and Tiny-ImageNet, achieving testing accuracies of 68.99\% and 64.44\%, respectively.
Together with improvements in training accuracy, these results demonstrate that our deep residual analytic architecture effectively enhances the fitting capability of ReCALM and leads to substantial performance gains.

To evaluate the effectiveness of our learned representations, we compare the intermediate features of ReCALM with those of DeepAFL~\cite{DeepAFL}, a deep analytic learning method.
Although DeepAFL supports distributed deep analytic learning, it cannot directly accommodate continual multi-layer updates when historical data are unavailable.
Therefore, we compare them under a centralized setting to investigate the effectiveness of their learned deep representations.
Specifically, we employ three feature-separability metrics, namely Compactness-Separation Ratio (CSR), Inverse Fisher Score (IFS), and Discriminative Measure (DM), where lower values indicate better feature separability.
The Appendix reports the corresponding results on the training and testing sets of CIFAR-100 at different network depths.
As shown, ReCALM consistently outperforms DeepAFL across all three metrics at every evaluated positive depth.
For instance, at twenty layers, ReCALM achieves lower testing CSR (1.5039 vs. 1.5804), IFS (3.0774 vs. 3.2096), and DM (3.0381 vs. 3.1927).
These results demonstrate that our residual analytic architecture learns more discriminative feature representations.

Furthermore, to evaluate our deepening strategy, we compare ReCALM with three deep analytic learning methods, including DeepAFL~\cite{DeepAFL}, Deep Stacking~\cite{DAN}, and Label Encoding~\cite{INS}.
Unlike ReCALM, these methods do not directly support continual multi-layer model construction when historical raw data are unavailable.
Therefore, we conduct these comparisons under a centralized setting, where the complete dataset and intermediate features are accessible to all methods.
At twenty layers, ReCALM achieves 67.88\% and 63.39\% testing accuracy on CIFAR-100 and Tiny-ImageNet, respectively, outperforming DeepAFL by 0.93\% and 1.05\%.
The advantages remain at fifty layers, where ReCALM reaches 68.99\% and 64.44\%, compared with 68.51\% and 63.74\% for DeepAFL.
The other two deepening strategies exhibit more limited or less consistent performance improvements as network depth increases.
The complete comparisons are provided in the Appendix, with trends shown in Fig.~\ref{app:fig:deepen-1}.
\textbf{\textit{Overall, our ReCALM effectively enhances model fitting and feature separability through deeper residual representations, leading to improved performance across the evaluated datasets.}}

\begin{table*}[t]
\centering
\caption{{ReCALM's accuracy and computation efficiency with different backbones on CIFAR-100.}} 
\renewcommand{\arraystretch}{1.2}
\resizebox{1.0\textwidth}{!}{
\begin{NiceTabular}{
l|
>{\centering\arraybackslash}p{1.3cm}>{\centering\arraybackslash}p{1.1cm}>{\centering\arraybackslash}p{1.2cm}|
>{\centering\arraybackslash}p{1.3cm}>{\centering\arraybackslash}p{1.1cm}>{\centering\arraybackslash}p{1.2cm}|
>{\centering\arraybackslash}p{1.3cm}>{\centering\arraybackslash}p{1.1cm}>{\centering\arraybackslash}p{1.2cm}
}
\toprule
\multicolumn{1}{c|}{\multirow{2.5}{*}{{Layers}}} 
& \multicolumn{3}{c|}{ResNet-18}
& \multicolumn{3}{c|}{ViT-B-16-I-1K}
& \multicolumn{3}{c}{ ViT-B-16-I-21K} \\
\cmidrule(lr){2-4}
\cmidrule(lr){5-7}
\cmidrule(lr){8-10}
& Test Acc. & Adv. & Cost
& Test Acc. & Adv. & Cost
& Test Acc. & Adv. & Cost \\
\midrule
AFCL & 58.56\% & $/$ & \; 99.56 s & 75.45\% & $/$ & 210.77 s & 86.35\% & $/$ & 210.14 s \\
ReCALM ($L=5$) & 65.25\% & +6.69\%  & 111.84 s & 78.40\% & +2.95\%  & 216.40 s & 88.27\% & +1.92\% & 223.15 s  \\
ReCALM ($L=10$) & 66.62\% & +8.06\%  & 118.42 s & 79.51\% & +4.06\%  & 224.20 s & 88.94\% & +2.59\% & 238.79 s  \\
ReCALM ($L=15$) & 67.34\% & +8.78\%  & 126.16 s & 80.27\% & +4.82\%  & 236.72 s & 89.25\% & +2.90\% & 239.56 s  \\
ReCALM ($L=20$) & 67.84\% & +9.28\%  & 139.62 s & 80.53\% & +5.08\%  & 248.71 s & 89.42\% & +3.07\% & 250.64 s  \\
ReCALM ($L=25$) & 68.41\% & +9.85\%  & 148.29 s & 80.93\% & +5.48\%  & 259.79 s & 89.56\% & +3.21\% & 259.90 s  \\
ReCALM ($L=30$) & 68.55\% & +9.99\% & 161.15 s & 81.15\% & +5.70\%  & 273.73 s & 89.61\% & +3.26\% & 272.81 s  \\
\bottomrule
\end{NiceTabular}
}

\label{app:tab:ViT}
\end{table*}

\begin{figure*}[t] 
\centering 

 \begin{minipage}[h]{0.49\linewidth}
		\centering
		\subfloat[Training Accuracy]{%
              \centering  
               \label{fig:Temporal-Invariance-CIFAR-100-migration-4}%
               \includegraphics[width=1.0\textwidth]{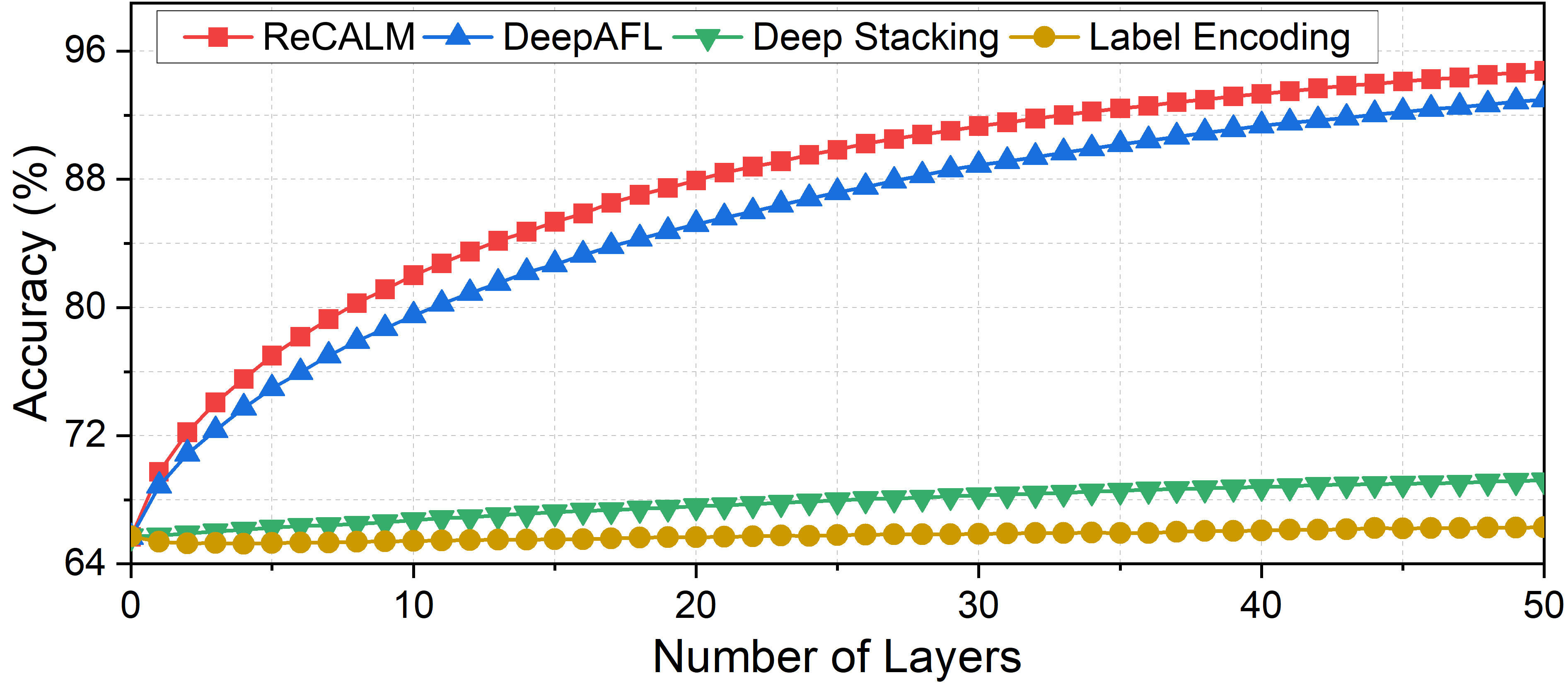}%
             } 
	\end{minipage}
	\hfill 
	\begin{minipage}[h]{0.49\linewidth}
		\centering
		\subfloat[Testing Accuracy]{%
              \centering  
               \label{fig:Temporal-Invariance-Tiny-ImageNet-free-migration-4}%
               \includegraphics[width=1.0\textwidth]{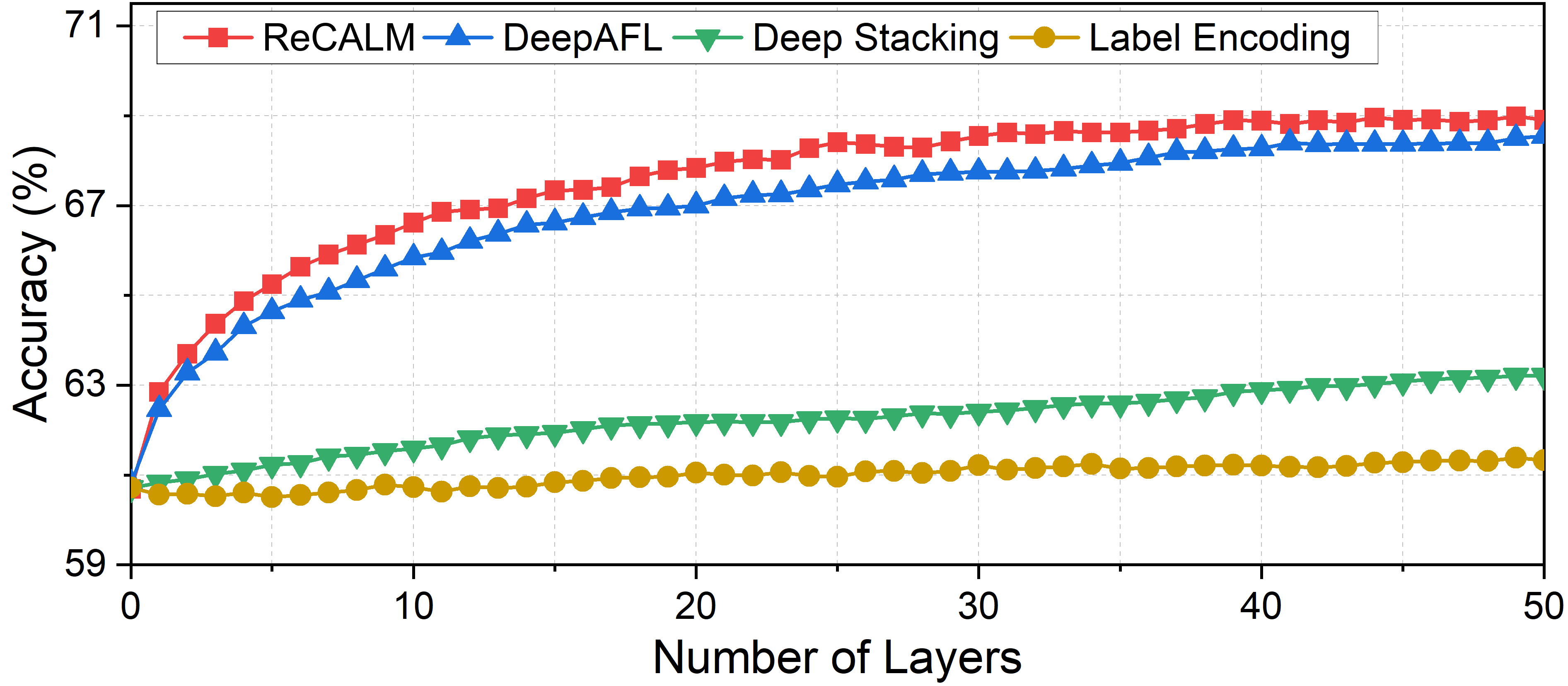}%
             } 
	\end{minipage}
    \vspace{-0.25cm}
 \caption{Comparison of our ReCALM with other deepening strategies on the CIFAR-100.
 } 
 \label{app:fig:deepen-1}
\end{figure*}

\begin{figure*}[t]
    \centering
    \includegraphics[width=1\linewidth]{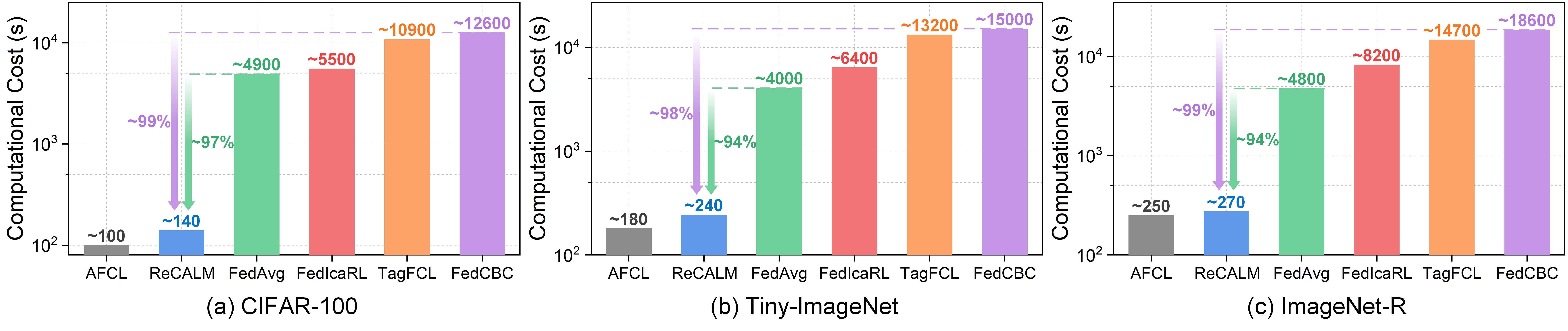}
     \vspace{-0.6cm}
    \caption{Efficiency comparisons between our ReCALM and the baselines across three datasets.}
    \label{fig:efficiency}
     \vspace{-0.2cm}
\end{figure*}

\subsection{RQ3: Efficiency Evaluation}
\label{sec:exp:rq3}

Here, we further evaluate the efficiency of ReCALM in crowdsourced continual service construction.
Specifically, Fig.~\ref{fig:efficiency} compares its computational cost with representative baselines across three datasets, while Table~\ref{app:tab:ViT} extends the evaluation to stronger backbones on CIFAR-100.
We also investigate how computational cost scales with network depth.

As shown in Fig.~\ref{fig:efficiency}, we compare the computational costs of our ReCALM with representative baselines across three datasets.
Considering that gradient-based methods generally incur similarly high computational costs, we select representative methods from FL (FedAvg), FL+CL (FedIcaRL), and FCL (TagFCL and FedCBC), together with the single-layer analytic method AFCL, for clarity.
Our ReCALM with twenty layers requires approximately 140, 240, and 270 seconds on CIFAR-100, Tiny-ImageNet, and ImageNet-R, respectively.
Compared with the gradient-based baselines, ReCALM reduces computational costs by 94\%--99\%, benefiting from its layer-wise closed-form computations without iterative back-propagation.
Meanwhile, compared with AFCL, our ReCALM incurs only a modest increase in computational cost while achieving substantial performance improvements through its additional residual layers.

Next, we further compare ReCALM with AFCL in Table~\ref{app:tab:ViT}, investigating the performance improvements and additional computational costs under different network depths and backbones.
As shown, ReCALM consistently achieves substantial performance gains with only modest additional computational overhead across all three backbones.
Specifically, at thirty layers, ReCALM outperforms AFCL by 9.99\%, 5.70\%, and 3.26\% with ResNet-18, ViT-B/16 (ImageNet-1K), and ViT-B/16 (ImageNet-21K), respectively, while requiring only about one additional minute of computation in each setting.
Although the performance gap gradually narrows as stronger backbones provide better initial features, ReCALM still achieves considerable improvements, highlighting the benefits of constructing deeper analytic networks even with powerful feature extractors.

In addition, we further investigate how the computational cost of ReCALM varies with network depth across all three datasets, with detailed results presented in the Appendix.
As shown, the computational cost increases gradually with network depth, with each additional residual layer requiring no more than three seconds.
Even when constructing a twenty-layer network, the total additional computation remains within a few tens of seconds, further demonstrating the efficiency of our deep residual analytic architecture.

\begin{table*}[t]
\centering
\caption{Accuracy of our ReCALM under varying dimensions  $d_{\Phi}$ and $d_\text{F}$.
For a comprehensive and consistent analysis, all experimental results were obtained by ensuring $d_{\Phi}= d_\text{F}=d$.
} 
\renewcommand{\arraystretch}{1.0}
\resizebox{1.0\textwidth}{!}{
\begin{NiceTabular}{@{}c|l|*{6}{ >{\centering\arraybackslash}m{17mm} }@{}}
\toprule[1.2pt]
\textbf{Datasets} &
\textbf{Layers}
& $d=2^{7}$ & $d=2^{8}$ & $d=2^{9}$ & $d=2^{10}$ & $d=2^{11}$ & $d=2^{12}$ \\
\midrule[0.4pt]
\multirow{3}{*}{CIFAR-100}
& $L=5$  & 56.66\% & 60.33\% & 62.97\% & 65.02\% & 67.24\% & \textbf{68.51\%}  \\
& $L=10$  & 59.05\% & 61.95\% & 64.46\% & 66.75\% & 68.48\% & \textbf{68.64\%}  \\
& $L=20$  & 60.78\% & 63.58\% & 65.87\% & 67.87\% & \textbf{69.09\%} & 68.45\%  \\

\midrule[0.4pt]
\multirow{3}{*}{Tiny-ImageNet}
& $L=5$  & 50.02\% & 56.49\% & 59.23\% & 61.05\% & 63.08\% & \textbf{64.05\%}  \\
& $L=10$  & 50.25\% & 58.10\% & 60.42\% & 62.23\% & 64.02\% & \textbf{64.61\%}  \\
& $L=20$  & 50.91\% & 59.52\% & 61.80\% & 63.52\% & 64.41\% & \textbf{64.65\%}  \\

\midrule[0.4pt]
    \multirow{3}{*}{ImageNet-R}
& $L=5$  & 35.72\% & 41.47\% & 44.52\% & 46.37\% & \textbf{47.50\%} & 45.90\%  \\
& $L=10$  & 37.95\% & 43.17\% & 46.13\% & 47.15\% & \textbf{47.47\%} & 44.82\% \\
& $L=20$  & 39.55\% & 44.98\% & 47.15\% & \textbf{48.05\% }& 46.87\% & 42.70\% \\

\bottomrule[1.2pt]
\end{NiceTabular}
}
\label{app:tab:d3}

\end{table*}

\begin{figure*}[t]
\begin{minipage}[h]{0.49\linewidth}
    \centering
    \subfloat[Training Accuracy]{%
          \centering  
           \label{fig:Temporal-Invariance-CIFAR-100}%
           \includegraphics[width=1.0\textwidth]{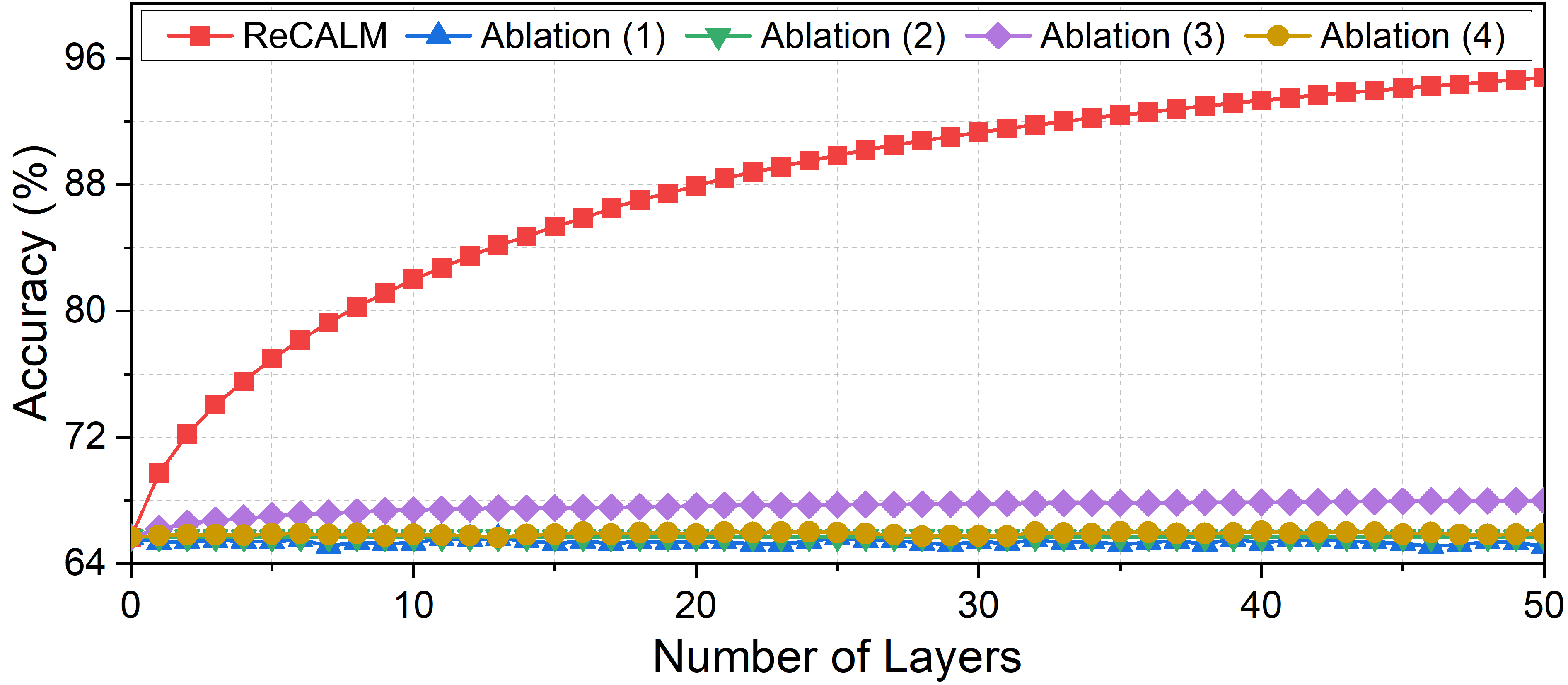}%
         } 
\end{minipage}
\hfill 
\begin{minipage}[h]{0.49\linewidth}
    \centering
    \subfloat[Testing Accuracy]{%
          \centering  
           \label{fig:Temporal-Invariance-Tiny-ImageNet-free}%
           \includegraphics[width=1.0\textwidth]{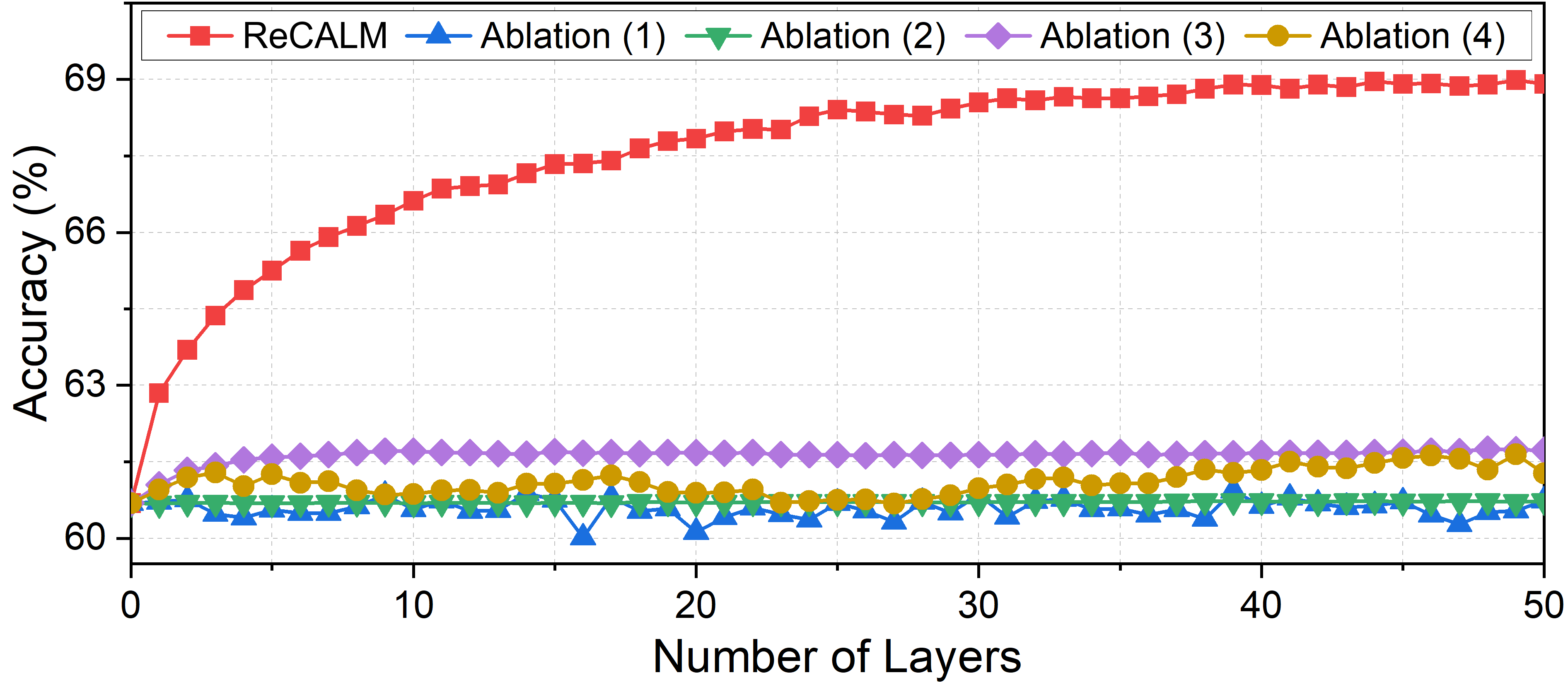}%
         } 
\end{minipage}
\vspace{-0.2cm}
\caption{Ablation study of our ReCALM on the CIFAR-100.} 
\vspace{-0.2cm}
\label{app:fig:Ablation-1}
\end{figure*}

\subsection{RQ4: Sensitivity and Ablation Studies}
\label{sec:exp:rq4}

Finally, we conduct comprehensive sensitivity analyses and ablation studies to investigate the effects of different parameter settings and architectural components on ReCALM's performance.
For the sensitivity analyses, we vary one parameter at a time while keeping all other parameters at their default settings.
Specifically, Table~\ref{app:tab:d3} and the accompanying results in the Appendix present detailed performance comparisons under different projection dimensions, regularization parameters, and activation functions.
Furthermore, Fig.~\ref{app:fig:Ablation-1}, along with the results in the Appendix, presents the ablation results across three datasets.
The detailed analyses are presented as follows.

As shown in Table~\ref{app:tab:d3}, jointly increasing the projection dimensions $d_\Phi$ and $d_{\text F}$ generally improves performance, particularly when the feature dimensions are relatively small.
At twenty layers, increasing both dimensions from 128 to 1024 improves accuracy from 60.78\% to 67.87\% on CIFAR-100, from 50.91\% to 63.52\% on Tiny-ImageNet, and from 39.55\% to 48.05\% on ImageNet-R.
Even with a dimension of 512, ReCALM achieves 65.87\%, 61.80\%, and 47.15\% on these datasets, respectively.
Further increasing the dimension to 2048 improves CIFAR-100 accuracy to 69.09\%, while ImageNet-R achieves its highest twenty-layer accuracy at a dimension of 1024.
Moreover, the individual analyses of $d_\Phi$ and $d_{\text F}$ in the Appendix further reveal the dataset-dependent effects of projection dimensions.
These results indicate that our default dimension of 1024 provides a favorable balance between performance and model complexity.

We further investigate the effects of regularization parameters and activation functions.
As shown in the Appendix, ReCALM exhibits stable performance across different regularization settings.
Specifically, varying $\lambda$ from 0.01 to 10 changes accuracy by at most 0.20\%, while the corresponding variation for $\gamma$ is at most 0.70\%.
Furthermore, the experimental results in the Appendix demonstrate that all six evaluated nonlinear activation functions outperform the setting without activation.
For instance, at twenty layers, GELU improves accuracy by 6.34\%, 5.96\%, and 6.42\% on CIFAR-100, Tiny-ImageNet, and ImageNet-R, respectively.
These results demonstrate the robustness of ReCALM to regularization choices and the importance of nonlinear activation in deep representation learning.

Next, we conduct ablation studies to investigate the contributions of four key components in \eqref{eq:section3.1-4}--\eqref{eq:section3.1-5}: the residual skip connection, random projection $\mathbf{B}_{l}$, activation function $\sigma(\cdot)$, and trainable transformation $\mathbf{\Omega}_{l+1}$.
As shown in Fig.~\ref{app:fig:Ablation-1}, the complete ReCALM achieves 87.93\% training accuracy and 67.88\% testing accuracy on CIFAR-100 at twenty layers.
Removing the skip connection, random projection, or trainable transformation reduces training accuracy to approximately 65\%--66\%, while testing accuracy decreases to 60.12\%, 60.70\%, and 60.89\%, respectively.
Similarly, removing the activation function reduces testing accuracy to 61.68\%.
The corresponding results on Tiny-ImageNet and ImageNet-R exhibit similar trends, as detailed in the Appendix.

These ablation results further demonstrate the importance of each component in our residual analytic architecture.
Specifically, the skip connection preserves existing representations, while the residual branch progressively introduces feature corrections.
The random projection and nonlinear activation enrich the feature representations, whereas the trainable transformation enables effective residual fitting.
\textbf{\textit{Overall, these results demonstrate that ReCALM achieves robust performance across different parameter settings, while its deep representation learning capability benefits from the combined contributions of these architectural components.}}

\section{Conclusion and Discussion}

In this paper, we propose ReCALM to explore deep nonlinear invariance to ST-DH in distributed continual learning. 
Our ReCALM inherits the gradient-free nature of analytic learning, transforming the costly iterative computations of backpropagation into recursive and forward-only updates. 
Our work extends traditional analytic learning from a single-layer network to a multi-layer residual network.
Our study can also be viewed as a non-trivial extension of DeepAFL~\cite{DeepAFL}, bringing the benefits of deep analytic networks from static FL to dynamic FCL. 
Specifically, we develop a residual analytic architecture and a novel cross-recursive strategy that combines phase-wise statistical accumulation with layer-wise reconstruction, enabling continual multi-layer learning without accessing historical raw data.
Extensive experiments demonstrate the effectiveness of ReCALM in improving performance while maintaining efficient model construction.
In future work, we plan to explore adaptive feature extraction within the continual analytic learning framework and further investigate the deployment of ReCALM in practical crowdsourcing systems with heterogeneous devices and dynamic client participation.

\bibliographystyle{ACM-Reference-Format}
\bibliography{11-paper}

\clearpage
\end{document}